%% file: root.tex
\documentclass[letterpaper, 10 pt, conference]{ieeeconf}  

\IEEEoverridecommandlockouts                              

\usepackage[english]{babel}
\usepackage{ifpdf}

\usepackage{cite} 
\usepackage{url}
\usepackage{hyperref}
\usepackage{xcolor}

\usepackage{color}
\usepackage{pgf, tikz, pgfplots}
\usetikzlibrary{shapes, arrows, automata}
\usetikzlibrary{calc,hobby,decorations}

\usepackage[cmex10]{amsmath}
\usepackage{amsfonts, amssymb}
\usepackage{mathrsfs}

\usepackage{algorithm} 
\usepackage{algorithmic} 
\usepackage{amsmath} 
\usepackage{xcolor}
\usepackage{pdfpages}
\usepackage{float}

\usepackage{enumerate}
\usepackage{multirow}
\usepackage{rotating}
\usepackage{subcaption}
\usepackage[labelfont=bf]{caption}

\input{mySections.tex}

\input{mySymbol.sty}

\definecolor{penndarkestblue}{cmyk}{1,0.74,0,0.77}
\definecolor{penndarkerblue}{cmyk}{1,0.74,0,0.70}
\definecolor{pennblue}{cmyk}{0.99,0.66,0,0.57} 
\definecolor{pennlighterblue}{cmyk}{0.98,0.44,0,0.35}
\definecolor{pennlightestblue}{cmyk}{0.38,0.17,0,0.17} 

\definecolor{penndarkestred}{cmyk}{0,1,0.89,0.66}
\definecolor{penndarkerred}{cmyk}{0,1,0.88,0.55}
\definecolor{pennred}{cmyk}{0,1,0.83,0.42} 
\definecolor{pennlighterred}{cmyk}{0,1,0.6,0.24}
\definecolor{pennlightestred}{cmyk}{0,0.43,0.26,0.12} 

\definecolor{penndarkestgreen}{cmyk}{1,0,1,0.68}
\definecolor{penndarkergreen}{cmyk}{1,0,1,0.57}
\definecolor{penngreen}{cmyk}{1,0,1,0.44} 
\definecolor{pennlightergreen}{cmyk}{1,0,1,0.25}
\definecolor{pennlightestgreen}{cmyk}{0.43,0,0.43,0.13}

\definecolor{penndarkestorange}{cmyk}{0,0.65,1,0.49}
\definecolor{penndarkerorange}{cmyk}{0,0.65,1,0.33}
\definecolor{pennorange}{cmyk}{0,0.54,1,0.24} 
\definecolor{pennlighterorange}{cmyk}{0,0.32,1,0.13}
\definecolor{pennlightestorange}{cmyk}{0,0.15,0.46,0.06}

\definecolor{penndarkestpurple}{cmyk}{0,1,0.11,0.86}
\definecolor{penndarkerpurple}{cmyk}{0,1,0.13,0.82}
\definecolor{pennpurple}{cmyk}{0,1,0.11,0.71} 
\definecolor{pennlighterpurple}{cmyk}{0,1,0.05,0.46}
\definecolor{pennlightestpurple}{cmyk}{0,0.35,0.02,0.23}

\definecolor{pennyellow}{cmyk}{0,0.20,1,0.05} 
\definecolor{pennlightgray1}{cmyk}{0,0,0,0.05}
\definecolor{pennlightgray3}{cmyk}{0.01,0.01,0,0.18}
\definecolor{pennmediumgray1}{cmyk}{0.04,0.03,0,0.31}
\definecolor{pennmediumgray4}{cmyk}{0.08,0.06,0,0.54}
\definecolor{penndarkgray2}{cmyk}{0.09,0.07,0,0.71}
\definecolor{penndarkgray4}{cmyk}{0.1,0.1,0,0.92}

\def\SO3{\mathrm{SO(3)}}

\newtheorem{problem}{\hspace{0pt}\bf Problem}

\newtheorem{definition}{\hspace{0pt}\bf Definition}

\title{\LARGE \bf
Online Control Barrier Functions \\ for Decentralized Multi-Agent Navigation 
}

\author{Zhan Gao, Guang Yang and Amanda Prorok
\thanks{Zhan Gao, Guang Yang and Amanda Prorok are with Department of Computer Science and Technology, University of Cambridge, CB3 0FD (email: zg292@cam.ac.uk; gy268@cam.ac.uk; asp45@cam.ac.uk).}
\thanks{
        This work was supported by ERC Project 949940 (gAIa).}
    }

\begin{document}

\maketitle
\thispagestyle{empty}
\pagestyle{empty}

\begin{abstract}
Control barrier functions (CBFs) enable \textit{guaranteed safe} multi-agent navigation in the continuous domain. The resulting navigation performance, however, is highly sensitive to the underlying hyperparameters. Traditional approaches consider \textit{fixed} CBFs (where parameters are tuned {apriori}), and hence, typically do not perform well in cluttered and highly dynamic environments: conservative parameter values can lead to inefficient agent trajectories, or even failure to reach goal positions, whereas aggressive parameter values can lead to infeasible controls. To overcome these issues, in this paper, we propose \textit{online CBFs}, whereby hyperparameters are tuned in real-time, as a function of what agents perceive in their immediate neighborhood. Since the explicit relationship between CBFs and navigation performance is hard to model, we leverage reinforcement learning to learn CBF-tuning policies in a model-free manner. Because we parameterize the policies with graph neural networks (GNNs), we are able to synthesize decentralized agent controllers that adjust parameter values locally, varying the degree of conservative and aggressive behaviors across agents. Simulations as well as real-world experiments show that \textit{(i)} online CBFs are capable of solving navigation scenarios that are infeasible for fixed CBFs, and \textit{(ii)}, that they improve navigation performance by adapting to other agents and changes in the environment.
\end{abstract}

\section{INTRODUCTION}

Multi-agent systems are ideally suited to tackle spatially distributed tasks, for which safe and efficient motion planning is a key enabling foundation \cite{ota2006multi, wang2017cooperative, oroojlooy2022review}. In the context of multi-agent navigation, model-based approaches often assume full knowledge of the environment and system dynamics, and require designing explicit objective functions and well-tuned hyper-parameters prior to agent deployment. Data-driven approaches are able to work with partially observed environments and complex system dynamics that are difficult to model, but often sacrifice safety and convergence guarantees. This work aims to find a middle-ground that leverages the advantages from both approaches.

In this paper, we focus on designing decentralized controllers for multi-agent navigation with dynamical constraints. Different from classic path-finding problems \cite{lavalle2001rapidly,wagner2011m,foead2021systematic}, we generate feedback control inputs and perform collision avoidance in continuous space. Specifically, the problem of multi-agent navigation with convergence and safety guarantees can be formulated as a sequence of real-time optimization problems by using control barrier functions (CBFs) and control Lyapunov functions (CLFs), where the former allow agents to move safely without collision and the latter guide them towards target states. The combination of CBFs and CLFs has been widely used for safety-critical controls \cite{ames2014control}, \cite{nguyen2016exponential}. Although CBFs provide safety guarantees, traditional approaches require manually setting parameters within CBF constraints \cite{hsu2015control, ames2014control, nguyen2016exponential, borrmann2015control}. This may yield overly conservative trajectories with strong CBF constraints or overly aggressive trajectories with relaxed ones, both of which could lead to controller infeasibility, i.e., no admissible control exists, and agents are hence unable to steer to their destinations. Moreover, such issues occur more frequently when the environment is cluttered with moving agents and an increasing number of obstacles, because the number of CBF constraints scales with that of agents and obstacles. Many existing works preset CBF parameters before deployment and fix the latter throughout the navigation procedure \cite{9303857, zeng2021enhancing, cheng2020safe}. This requires resetting CBF parameters in each new environment, and makes multi-agent systems incapable of operating in dynamic environments where agent configurations and obstacle constellations vary across time. Hence, we aim to develop methods that capture time-varying environment states and tune CBF parameters in real time as a function of these states.

Instead of hand-tuning and fixing CBF parameters at the outset, we propose a methodology that tunes the latter based on agent and obstacle states in a real-time and decentralized manner. The goal is to find an optimal sequence of time-varying CBF parameters that adapts to new environment configurations, and that varies the degree of conservative and aggressive behaviors across agents (striking a balance that aids in trajectory deconfliction). Due to the challenge of explicitly modeling the relationship between CBFs and navigation performance, we parameterize the CBF-tuning policy with graph neural networks (GNNs) and learn the latter with model-free reinforcement learning (RL). Thanks to the inherently distributed nature of GNNs \cite{gao2022environment, li2020graph, tolstaya2020learning, gao2022wide}, the resulting policy allows for a decentralized implementation, i.e., it can be executed by each agent locally with only neighborhood information, yielding an efficient and scalable solution. 

\textbf{Related work.} There are two main groups of CBF-based techniques in multi-agent control: model-based \cite{lindemann2019control,tan2021distributed,9304151,srinivasan2018control} and data-driven approaches \cite{qin2021learning,ahmadi2019safe,yu2022learning}. Model-based approaches require full knowledge of the environment and fix CBF parameters a-priori. The CBF constraints are affine in the control variable, to formulate a quadratic program (QP) controller, which provides safety guarantees for navigation. In contrast, data-driven approaches directly approximate CBFs with neural networks \cite{dawson2022safe, robey2020learning, pauli2021training}. However, these approaches sacrifice safety guarantees in the process. The work in \cite{9300218} combines model-based and data-driven approaches by learning a backup CBF to enhance the safety. For the feasibility of CBFs, \cite{xu2022feasibility} designs a feasibility guaranteed controller for traffic-merging problems. The work in \cite{9483029} studies the feasibility of a CBF-based model predictive controller (MPC) in a discrete time setting, while \cite{zeng2021enhancing} introduces a decaying term paired with CBFs to improve the feasibility of the MPC. Moreover, \cite{breeden2021control} extends the class $\mathcal{K}$ function of CBFs for forward invariance in continuous time. A more closely related work, \cite{9303857}, develops an SVM classifier to filter out infeasible CBF parameters and reduce the search space to find optimal CBF parameters. However, it considers single-agent scenarios and the selected CBF parameters are fixed during navigation. To the best of our knowledge, \textit{none of the aforementioned works update CBF parameters online based on changes in the agents' locally perceived environment.} 

\textbf{Contributions.} Our contributions are as follows: 
\begin{enumerate}
  \item We propose an \textit{online} safety-critical framework that adapts CBFs to dynamic environments in a decentralized manner. It inherits safety guarantees from traditional CBFs and facilitates the feasibility of the controller, due to the online tuning of CBF parameters. 

  \item We parameterize the CBF-tuning policy with GNNs and conduct training with model-free RL. The former allows for a decentralized implementation, while the latter overcomes the challenge of explicitly modeling the relationship between CBFs and navigation performance. 
  
  \item We validate our approach with numerical simulations and real-world experiments in various environment configurations. The results show that online CBFs can handle navigation scenarios that fail with fixed CBFs (even when we perform an exhaustive parameter search).  
\end{enumerate}

\section{PRELIMINARIES}

We introduce preliminaries about system dynamics, CLFs and CBFs in decentralized multi-agent navigation.

\begin{figure}[tb]
\centering
\includegraphics[width=0.25\textwidth]{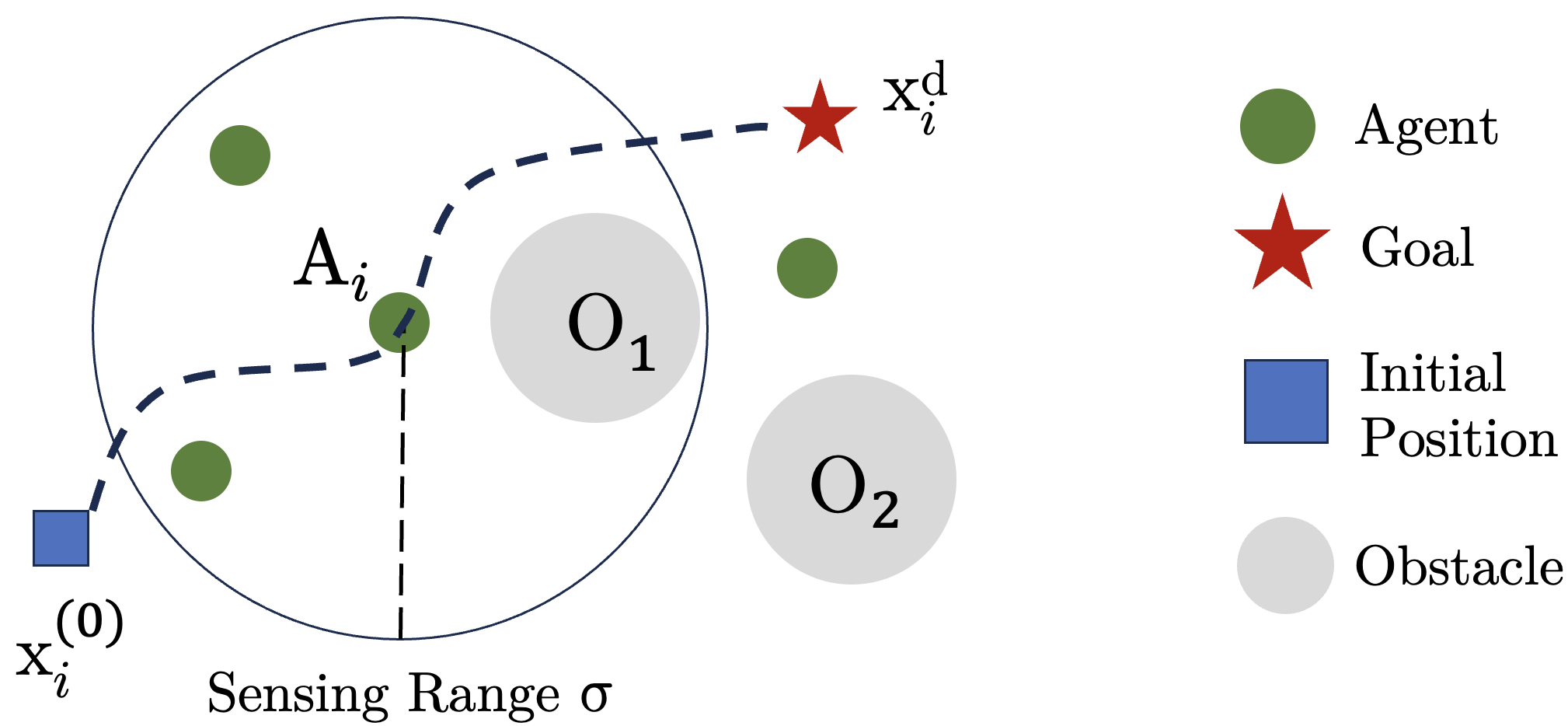}
\captionof{figure}{Agent $A_i$ communicates with the other agents and senses obstacles within its sensing range $\sigma$. In this sketch, there are two CBF constraints w.r.t. the other agents, and one w.r.t. obstacle $O_1$. 
}
\label{fig:U_CBF_CLF}\vspace{-6mm}
\end{figure}

\noindent \textbf{System dynamics}
Consider a multi-agent system with $N$ agents $\ccalA = \{A_i\}_{i=1}^N$ in a 2-D environment with $M$ static obstacles $\{O_j\}_{j=1}^M$. The agent dynamics take the form of 
\begin{equation}\label{eq:dynamicSystem}
\dot{\bbx}_i = f(\bbx_i) + g(\bbx_i)\bbu_i,
\end{equation} 
where $\bbx_i \in \mathbb{R}^n$ is the internal state, $\bbu_i \in \mathbb{R}^m$ the control input, $\dot{\bbx}_i$ the derivative of $\bbx_i$ w.r.t. time $t$, and $f(\bbx_i)$, $g(\bbx_i)$ the flow vectors for $i=1,...,N$. Each agent $A_i$ has a sensing range $\sigma \in \mathbb{R}^+$ that provides 
partial observability of the entire environment, i.e., 
the states of the other agents $\{\bbx_j\}_{j \in \ccalN_i}$ and the positions of the obstacles $\{\bbp_{\ell,o}\}_{\ell \in \ccalN_i}$ 
within the neighborhood of radius $\sigma$ where $\ccalN_i$ is the neighbor set of $A_i$ 
-- see Fig. \ref{fig:U_CBF_CLF}. 
We consider \emph{decentralized} control policies 
\begin{equation}\label{eq:decentralizedControlPolicy}
    \pi_{i} \Big(\bbu_i \Big| \bbx_i, \{\bbx_j\}_{j \in \ccalN_i}, \{\bbp_{\ell,o}\}_{\ell \in \ccalN_i}\Big),~\text{for}~i=1,\ldots,N 
\end{equation}
that drive agents from initial 
$\bbX^{(0)} := \{\bbx_{i}^{(0)}\}_{i=1}^N$ to target states $\bbX^{d} := \{\bbx_{i}^{d}\}_{i=1}^N$ with local neighborhood information.  

\noindent \textbf{Control Lyapunov function (CLF).}
A CLF is designed to encode the goal-reaching requirement, i.e., the satisfaction of CLF constraints 
guarantees that agents converge to their target states. We define the exponentially-stabilizing CLF that ensures an exponential convergence as follows 
\cite{ames2014rapidly}.
%

\begin{definition}
Given the system dynamics \eqref{eq:dynamicSystem} of agent $A_i$, a positive definite continuously differentiable function $V_i(\bbx_i): \mathbb{R}^n \mapsto \mathbb{R}$ is an exponentially-stabilizing CLF if there exists a positive constant $\epsilon \geq 0$ such that for any $\bbx_i \!\in\! \mathbb{R}^n$, 
\begin{align}\label{eq:CLF}
\!\textstyle \inf_{\bbu_i\in \mathbb{U}_i} [\pounds_{f} V_i(\bbx_i\!)\!+\!\pounds_{g}V_i(\bbx_i)\bbu_i\!+\!\epsilon V_i(\bbx_i)] \!\leq\! 0,  
\end{align}
where $\pounds_{f} V_i(\bbx_i):= \frac{\partial V_i(\bbx_i)}{\partial \bbx_i} f(\bbx_i)$ is the Lie derivative of 
$V_i(\bbx_i)$ 
\cite{khalil2002nonlinear} 
and $\mathbb{U}_i$ is the control space of agent $A_i$.
\end{definition}

\noindent \textbf{Control barrier function (CBF).}
A CBF is designed to avoid static obstacles as well as to prevent collisions among moving agents. It ensures \textit{forward invariance} of the state trajectory, i.e., if the agent starts within a safety set, it will always stay within safety sets \cite{xu2015robustness}. Specifically, we encode the safety requirement of agent $A_i$ in a smooth function $h_i(\bbx_i):\mathbb{R}^n \mapsto \mathbb{R}$ and its derivative w.r.t. time is given by 
\begin{equation} \label{eq:relative_degree}
\dot{h}_i(\bbx_i) = \pounds_{f} h_i(\bbx_i)+\pounds_{g}h_i(\bbx_i)\bbu_i,
\end{equation}
where $\pounds_{f} h_i(\bbx_i) := \frac{\partial  h_i (\bbx_i)}{\partial \bbx_i} f(\bbx_i), \pounds_{g} h_i(\bbx_i) := \frac{\partial h_i (\bbx_i)}{\partial \bbx_i} g(\bbx_i)$ are the Lie derivatives of $ h_i (\bbx_i)$. 
We define the higher-order CBF with relative degree one as follows \cite{xiao2019control}. 
\begin{definition}
Given the system dynamics \eqref{eq:dynamicSystem} of agent $A_i$, a differentiable function $h_i:\mathbb{R}^n \mapsto \mathbb{R}$ is a higher-order CBF with relative degree one if there exists a strictly increasing function $\alpha_i: \mathbb{R}^+ \to \mathbb{R}^+$ such that for any $\bbx_i \!\in\! \mathbb{R}^n$, 
\begin{align}\label{eq:CBFConstraints}
    h_i(\bbx_i) \ge 0,~ \dot{h}_i(\bbx_i) +  \alpha_{i}\big(h_i(\bbx_i)\big) \ge 0.
\end{align}
Here, $\alpha_i(\cdot)$ is referred to as a class $\mathcal{K}$ function for agent $A_i$, which is determined by function parameters 
$\eta$ and $\zeta$. More details are given in Section \ref{subsec:CLF-CBF-QP}. 
\end{definition}

\section{PROBLEM FORMULATION}

We first formulate the problem of decentralized multi-agent navigation with CLFs for state convergence and CBFs for safety guarantees. We then propose the problem of online CBF optimization, which generates time-varying CBFs based on instantaneously sensed states to optimize performance. 

\subsection{Decentralized Multi-Agent Navigation}

Assume agents and obstacles are disk-shaped with radii $\{R_i\}_{i=1}^N$ and $\{R_\ell\}_{\ell=1}^M$. 
Let $\{\bbp_i\}_{i=1}^N$, $\{\bbv_i\}_{i=1}^N$ and $\{\bbd_i\}_{i=1}^N$ be the positions, velocities and destinations of agents $\ccalA$, which are determined by the internal states $\{\bbx_i\}_{i=1}^N$, 
and $\{\bbp_{\ell,o}\}_{\ell=1}^M$ be the obstacle positions. 
The goal 
is to move agents towards destinations while avoiding collision in a decentralized manner. The destination convergence is equivalent to the state convergence as $\lim_{t \to T} \bbx^{(t)}_i = \bbx^d_i$ with $T$ the maximal time step for $i=1,...,N$. The collision avoidance is equivalent to the safety constraints on the agent states, i.e.,
\begin{align}\label{eq:safety1}
    &\ccalC^{(t)}_{i,\rm a} \!=\!\! \{\bbx^{(t)}_i \!\in\! \mathbb{R}^n \!\!~|~\! \| \bbp^{(t)}_i \!-\! \bbp^{(t)}_j \| \!\geq\! \frac{R_i \!\!+\!\! R_j}{2},~j \!\ne\! i\}, \\
    \label{eq:safety2}&\ccalC^{(t)}_{i,\rm o} \!=\!\! \{\bbx^{(t)}_i \!\in\! \mathbb{R}^n \!\!~|~\! \| \bbp^{(t)}_i \!-\! \bbp^{(t)}_{\ell,\rm o}\| \!\!\geq\!\! \frac{R_i \!\!+\!\! R_\ell}{2}\!,~\!\ell\!=\!1,...,M\}. 
\end{align}
where $\|\cdot\|$ is the vector norm. This allows us to formulate the problem of multi-agent navigation as follows.
\begin{problem}[Decentralized Multi-Agent Navigation]\label{def:problem1}
Given the multi-agent system $\ccalA$ with dynamics \eqref{eq:dynamicSystem}, the initial states $\{\bbx_{i}^{(0)}\}_{i=1}^{N
}$ and the target states $\{\bbx_{i}^d\}_{i=1}^N$ satisfying $\{f(\bbx_{i}^d) = 0\}_{i=1}^N$, find decentralized policies $\{\pi_i\!:\! \mathbb{R}^{|\ccalN_i| \times n} \!\to\! \mathbb{R}^m\}_{i=1}^N$ [cf. \eqref{eq:decentralizedControlPolicy}] 
such that for $i=1,...,N$, \vspace{-1mm}
\begin{align}\label{eq:convergence}
	&\lim_{t \to T}~~ \|\bbx_i^{(t)} - \bbx_{i}^d\| = 0,\\
	\label{eq:safety}&~~\text{s.t.}~~ \bbx_i^{(t)} \in \ccalC^{(t)}_{i,\textrm{a}} \cap \ccalC^{(t)}_{i,\textrm{o}} = \ccalC^{(t)}_{i}. 
\end{align} 
\end{problem}

\smallskip
The condition \eqref{eq:convergence} guarantees state convergence, i.e., navigation, and the condition \eqref{eq:safety} guarantees state safety, i.e., collision avoidance. Problem \ref{def:problem1} is challenging because (\textit{i}) decentralized policies 
generate control inputs with only local neighborhood information; and (\textit{ii}), safety sets can be non-convex and time-varying, depending on moving agents. 

We propose to solve Problem \ref{def:problem1} with a CBF-CLF based quadratic programming (QP) controller. Specifically, at each time step $t$, we can formulate a QP problem as \vspace{-1mm}
\begin{align}\label{eq:ECBF-ZCBF-QP}
& ~\underset{\bbu^{(t)}_i\in \mathbb{U}_{i},\delta_i^{(t)} \in \mathbb{R}}{\text{min}}
& & \| \bbu_i^{(t)}\|_{2} + \xi (\delta_i^{(t)})^2 \\
& ~\text{s.t.}
& & \!\!\!\!\!\!\!\!\!\!\!\!\!\!\!\!\!\!\!\!\!\!\!\!\!\!\!\!\pounds_{f} h_{i,j}(\bbx_i^{(t)})\!+\!\pounds_{g} h_{i,j}(\bbx_i^{(t)})\bbu_i^{(t)}\!\!+\! \alpha_{i}(h_{i,j}(\bbx_i^{(t)})) \!\geq\! 0,\nonumber\\
& & &\!\!\!\!\!\!\!\!\!\!\!\!\!\!\!\!\!\!\!\!\!\!\!\!\!\!\!\!\text{for all}~j=0,\dots,C_{i},\nonumber\\
& & &\!\!\!\!\!\!\!\!\!\!\!\!\!\!\!\!\!\!\!\!\!\!\!\!\!\!\!\!\pounds_{f} V_i(\bbx_i^{(t)})\!+\!\pounds_{g}V_i(\bbx_i^{(t)})\bbu_i^{(t)}\!+\!\epsilon V_i(\bbx_i^{(t)}) \!+\! \delta_i^{(t)} \leq 0,\nonumber\\
& & &\!\!\!\!\!\!\!\!\!\!\!\!\!\!\!\!\!\!\!\!\!\!\!\!\!\!\!\!\bbx_i^{(t)}\in \ccalC_i^{(t)},~\text{for all}~i=1,\dots,N, \nonumber 
\end{align}
where $\xi \in \mathbb{R}^+$ is a penalty weight for slack variable $\delta_i \in \mathbb{R}$ that is selected based on how strictly the CLF needs to be enforced, 
and $C_{i}$ is the number of CBF constraints based on agent $A_i$'s perceived agents and obstacles. For example, there are two CBF constraints w.r.t. the other agents and one w.r.t. the obstacle in Fig. \ref{fig:U_CBF_CLF}. The control input is bounded by $\bbu_i^{(t)} \in \mathbb{U}_{i}$ given physical constraints of agent $A_i$. 
The QP is solved per time step to generate $\bbu_i^{(t)}$ until completion. 
The 
class $\mathcal{K}$ function $\alpha_{i}(\cdot)$ in the CBF constraint determines how strictly we want to enforce safety, and therefore will change the agent behavior 
to be either conservative or aggressive. 

\subsection{Online CBF Optimization} \label{subsec:CBFProblem}
The CBF-CLF-QP controller solves problem \eqref{eq:ECBF-ZCBF-QP} to generate a sequence of control inputs $\{\bbu_i^{(t)}\}_{t=0}^{T}$ for each agent $A_i$, providing goal-reaching convergence and safety guarantees. However, this may not be the case when problem \eqref{eq:ECBF-ZCBF-QP} is unsolvable at some time step $t$ during navigation, i.e., 
when there is no feasible solution for problem \eqref{eq:ECBF-ZCBF-QP} given CBF and CLF constraints at time step $t$. In this circumstance, an agent will stay safe at its current state but stop progressing towards its destination due to the lack of feasible control inputs, resulting in the failure of multi-agent navigation. 

Specifically, 
given system dynamics \eqref{eq:dynamicSystem}, CBF constraints \eqref{eq:CBFConstraints} and CLF constraints \eqref{eq:CLF}, the super-level set of agent $A_i$ that satisfies the constraints in problem \eqref{eq:ECBF-ZCBF-QP} at time step $t$ is \vspace{-1mm}
\begin{equation} \label{eq:CBFControlset}
\mathbb{U}^{(t)}_{i,\mathrm{CBF},\mathrm{CLF}} \!\!:=\!\! \left\{ 
\!\!\bbu_i^{(t)} \!\!\!\;\middle|\;\!\!
\begin{aligned}
       & \!\pounds_{g} h_{i,j}(\bbx_i^{(t)})\bbu_i^{(t)} \!\!\geq\!\! -\pounds_{f} h_{i,j}(\bbx_i^{(t)}\!)\\&\!-\!\alpha_{i} \big(h_{i,j}(\bbx_i^{(t)})\big),\text{for all}~j,
       \\ & -\pounds_{g} V_i(\bbx_i^{(t)})\bbu_i^{(t)} \!\!\geq\! \pounds_{f} V_i (\bbx_i^{(t)}) \\&+ \epsilon V_i (\bbx_i^{(t)})+\delta_i^{(t)}
\end{aligned}
\!\!\right\}\!\!.
\end{equation}
By combining \eqref{eq:CBFControlset} with the physical constraints $\mathbb{U}_{i}$, the space of feasible solutions of agent $A_i$ is given by 
\begin{equation} \label{eq:feasible_condtion}
    \mathbb{U}_{i} \cap \mathbb{U}^{(t)}_{i, \mathrm{CBF},\mathrm{CLF}},~\for~i=1,\ldots,N. 
\end{equation}
For conservative CBFs, the resulting constraints are strict and there may be no feasible solution in $\mathbb{U}^{(t)}_{i,\mathrm{CBF},\mathrm{CLF}}$ s.t. $\mathbb{U}_{i} \cap \mathbb{U}^{(t)}_{i, \mathrm{CBF},\mathrm{CLF}} = \varnothing$. For aggressive CBFs, the resulting constraints are relaxed and the agents may be too close to the obstacles or each other. In these cases, the QP controller may generate control inputs that require sudden changes beyond the agent's physical capability $\mathbb{U}_{i}$ s.t. 
$\mathbb{U}_{i} \cap \mathbb{U}^{(t)}_{i, \mathrm{CBF},\mathrm{CLF}} = \varnothing$. Both scenarios lead to the infeasibility of problem \eqref{eq:ECBF-ZCBF-QP} and, thus, navigation failure, 
indicating an inherent trade-off that is defined by CBF constraints -- see Figs. \ref{subfig3a}-\ref{subfig3b} for examples.

The aforementioned issue is exacerbated when the environment becomes cluttered with increasing numbers of agents and obstacles, which makes it challenging to hand-tune CBFs. Furthermore, fixing CBFs during navigation may not effectively handle the dynamic nature of the environment with moving agents, and even well-tuned CBFs could suffer from performance degradation with environment changes. These observations motivate the use of time-varying CBFs based on instantaneously sensed states, to tune agents' conservative and aggressive behavior. We refer to the latter as \textit{online CBFs}. Define decentralized CBF-tuning policies as 
\begin{equation} \label{eq:cbf_policy}
    \pi_{i, \mathrm{CBF}}\Big(\!\alpha_{i}\Big| \bbx_i,\! \{\bbx_j\}_{j \in \ccalN_i},\! \{\bbp_{\ell,\rm o}\}_{\ell \in \ccalN_i}\!\Big),~\for~i\!=\!1,...,N,
\end{equation}
which generate the class $\ccalK$ function $\alpha_{i}(\cdot)$, i.e., the CBFs, based on local neighborhood information.  
At each time step $t$, a new class $\ccalK$ function $\alpha^{(t)}_{i}(\cdot)$ is generated for agent $A_i$ and passed into CBFs for solving \eqref{eq:ECBF-ZCBF-QP} to compute the control input $\bbu_i^{(t)}$. Given any objective function $F(\{\pi_{i,\mathrm{CBF}}\}_{i=1}^N, \bbX^{(0)}, \bbX^d)$ that represents the navigation performance, 
the initial states $\bbX^{(0)}$ and the target states $\bbX^d$, we can formulate the problem of online CBF optimization. 
\begin{problem}[Online CBF optimization]\label{problem2}
Given the 
initial states $\bbX^{(0)}$ and target states $\bbX^d$, find decentralized CBF-tuning policies $\{\pi_{i,\mathrm{CBF}}\}_{i=1}^N$ [cf. \eqref{eq:cbf_policy}] that generate online CBFs with local neighborhood information, to guarantee the feasibility of problem \eqref{eq:ECBF-ZCBF-QP} and maximize the objective function $F(\{\pi_{i,\mathrm{CBF}}\}_{i=1}^N, \bbX^{(0)}, \bbX^d)$. 
\end{problem}

The CBF-tuning policy conducts online CBF adjustments 
based on the local state of a dynamic environment, which provides control feasibility where fixed CBFs would not be able to, while maintaining safety guarantees. The generated time-varying CBFs strike a balance between conservative and aggressive behaviors among different agents. For scenarios where agent trajectories are in conflict (e.g., several agents need to navigate through narrow space), this yields an inherent prioritization among agents and provides deconfliction for agent trajectories -- see Fig. \ref{subfig4a} for demonstration. 

\section{METHODOLOGY}\label{sec:methodology}
In this section, we specify the decentralized CLF-CBF-QP controller to solve Problem \ref{def:problem1} and leverage model-free reinforcement learning with decentralized GNNs to solve Problem \ref{problem2}. We consider a linear system for each agent $A_i$, which has the following system dynamics 
\begin{align} \label{eq:linearsystemdynamics}
\left[\begin{matrix}\dot{p}_{i,1} \\ \dot{p}_{i,2} \end{matrix} \right]= \left[ \begin{matrix} 0&0\\ 0&0 \end{matrix} \right] \left[\begin{matrix} p_{i,1} \\ p_{i,2} \end{matrix} \right]+ \left[ \begin{matrix} 1&0\\ 0&1 \end{matrix} \right] \left[ \begin{matrix} u_{i,1} \\ u_{i,2}\end{matrix} \right],
\end{align}
where $\bbp_i = [p_{i,1}, p_{i,2}]^\top$ is the position and $\bbu_i = [u_{i,1},u_{i,2}]^\top$ is the control input of agent $A_i$ for $i=1,...,N$. 

\subsection{CLF-CBF-QP Controller}\label{subsec:CLF-CBF-QP}
Given the destination $\bbd_{i}=[d_{i, 1}, d_{i,2}]^\top$ of agent $A_i$, define a Lyapunov function candidate as $V_i(\bbx) = (p_{i,1}-d_{i, 1})^2+(p_{i,2}-d_{i,2})^2$ and the CLF constraint as 
\begin{equation}\label{eq:CLF_robots}
2(\bbp_{i}-\bbd_{i})^\top\bbu_{i} + \epsilon V_i(\bbx_i) + \delta_i \leq 0
\end{equation}
for $i = 1,...,N$. Define barrier function candidates as
\begin{align} \label{eq:CBF_robots_obs}
h_{i,j,\mathrm{a}}(\bbx_i) &\!=\! (p_{i,1} \!-\! p_{j,1})^2 \!+\! (p_{i,2} \!-\! p_{j,2})^2 \!-\! (R_i\!+\!R_j)^2, \\ \nonumber
h_{i,\ell,\mathrm{o}}(\bbx_i) &\!=\! (p_{i,1} \!-\! p_{\ell,1,\mathrm{o}})^2 \!+\! (p_{i,2} \!-\! p_{\ell,2,\mathrm{o}})^2 \!-\! (R_i\!+\!R_\ell)^2, 
\end{align}
where $h_{i,j,\mathrm{a}}$ is w.r.t. collision avoidance between the agents $A_i$ and $A_j$, and $h_{i,\ell,\mathrm{o}}$ is w.r.t. collision avoidance between agent $A_i$ and obstacle $O_\ell$ with $\bbp_{\ell,\mathrm{o}}=[p_{\ell,\mathrm{1, o}}, p_{\ell,\mathrm{2, o}}]^\top$ the obstacle position. 
The resulting CBF constraints are 
\begin{align}\label{eq:nu_CBF}
2(\bbp_{i}-&\bbp_{j})^\top \bbu_i-2(\bbp_{i}-\bbp_{j})^\top \dot{\bbp}_{j} + \zeta_{i,\mathrm{a}} \big(h_{i,j,\mathrm{a}}(\bbx_i)\big)^{\eta_{i,\mathrm{a}}} \geq 0,  \nonumber\\
&2(\bbp_{i}-\bbp_{\ell})^\top\bbu_i + \zeta_{i,\mathrm{o}} \big(h_{i,\ell,\mathrm{o}}(\bbx_i)\big)^{\eta_{i,\mathrm{o}}} \geq 0,
\end{align}
where $\zeta_{i,\mathrm{a}}, \eta_{i,\mathrm{a}}$ are CBF parameters of agent $A_i$ w.r.t. the other agents and $\zeta_{i,\mathrm{o}}, \eta_{i,\mathrm{o}}$ are 
w.r.t. the obstacles. In this context, each agent has two sets of CBF parameters for the other agents and for the obstacles, respectively. We can then specify the CBF-CLF-QP controller 
by substituting the CLF constraint \eqref{eq:CLF_robots} and CBF constraints \eqref{eq:nu_CBF} into problem \eqref{eq:ECBF-ZCBF-QP}. 

\subsection{Reinforcement Learning}
The CBFs 
are determined by the class $\ccalK$ function $\alpha_{i}(\cdot)$ with parameters $\zeta_{i,\mathrm{a}}, \eta_{i,\mathrm{a}}$ for the other agents and $\zeta_{i,\mathrm{o}}, \eta_{i,\mathrm{o}}$ for the obstacles. This indicates that we can learn CBFs by learning CBF parameters $\bbzeta_i = [\zeta_{i,\mathrm{a}}, \zeta_{i,\mathrm{o}}]^\top, \bbeta_i =[\eta_{i,\mathrm{a}}, \eta_{i,\mathrm{o}}]^\top$ at each agent $A_i$. 
Since it is challenging to explicitly model the relationship between CBF parameters and navigation performance, we formulate Problem \ref{problem2} in the RL domain and learn CBF-tuning policies in a model-free manner. 

We start by defining a partially observable Markov decision process. At each time $t$, agents are defined by states $\bbX^{(t)} \!=\! \{\bbx_i^{(t)}\}_{i=1}^N$. Each agent $A_i$ observes its local state $\bbx_i^{(t)}$, communicates with its neighboring agents, and senses its neighboring obstacles to collect the neighborhood information $\{\bbx_j^{(t)}\}_{j \in \ccalN_i}$ and $\{\bbp_{\ell,\mathrm{o}}\}_{\ell \in \ccalN_i}$.
The CBF-tuning policy $\pi_{i, \rm CBF}$ generates CBF parameters $\bbzeta_i^{(t)}$ and $\bbeta_i^{(t)}$, 
which is a distribution over $\bbzeta_i^{(t)}$, $\bbeta_i^{(t)}$ conditioned on $\bbx_i^{(t)}$, $\{\bbx_j^{(t)}\}_{j \in \ccalN_i}$, $\{\bbp_{\ell,\mathrm{o}}\}_{\ell \in \ccalN_i}$. 
The CBF parameters $\bbzeta_i^{(t)}$, $\bbeta_i^{(t)}$ are fed into the QP controller \eqref{eq:ECBF-ZCBF-QP}, 
which generates the control action $\bbu_i^{(t)}$ that drives the local state $\bbx_i^{(t)}$ to $\bbx_i^{(t+1)}$ based on the agent's dynamics \eqref{eq:linearsystemdynamics}. 
The reward function $r_i(\bbX^{(t)})$ represents the instantaneous navigation performance of agent $A_i$ at time $t$, which consists of two components: (\textit{i}) the navigation reward $r_{i,\textrm{nav}}$ and (\textit{ii}) the QP's feasibility reward $r_{i,\textrm{infs}}$, i.e., 
\begin{align}\label{eq:reward}
    r_i^{(t)}(\bbX^{(t)}) = r_{i, \text{nav}}^{(t)}(\bbX^{(t)}) + \beta_i r_{i, \text{infs}}^{(t)}(\bbX^{(t)}),
\end{align}
where 
$\beta_i$ is the regularization parameter. The first term 
represents the task-relevant performance of agent $A_i$, while the second term 
corresponds to the feasibility of the QP controller \eqref{eq:ECBF-ZCBF-QP} with the generated CBF parameters, e.g., it penalizes the scenario where the 
QP controller has no feasible solution with overly conservative or aggressive CBFs. 
The total reward of agents 
is $r^{(t)} = \sum_{i=1}^N r_i^{(t)}$. With the discount factor $\gamma$ that accounts for the future rewards, the expected discounted reward can be represented as 
\begin{align}\label{eq:expectedDiscounted}
    &R\big(\bbX^{(0)}\!, \bbX^{d}\!, \{\bbp_{\ell,\mathrm{o}}\}_{\ell=1}^M | \{\pi_{i, \rm CBF}\}_{i=1}^N\big) \!=\! \mathbb{E}\Big[\!\sum_{t=0}^\infty \gamma^t r^{(t)}\!\Big]\!,
\end{align}
where $\mathbb{E}[\cdot]$ is w.r.t. CBF-tuning policies. The expected discounted reward in \eqref{eq:expectedDiscounted} corresponds to the objective function in Problem \ref{problem2}, which transforms the problem 
into the RL domain. By parameterizing the policies $\{\pi_{i, \rm CBF}\}_{i=1}^N$ with information processing architectures $\{\bbPhi_i(\bbx_i^{(t)}, \{\bbx_j^{(t)}\}_{j \in \ccalN_i},\{\bbp_{\ell,\mathrm{o}}\}_{\ell \in \ccalN_i}, \bbtheta_i)\}_{i=1}^N$ of parameters $\{\bbtheta_i\}_{i=1}^{N}$, the goal is to learn optimal parameters $\{\bbtheta_i^*\}_{i=1}^{N}$ that maximize 
$R(\bbX^{(0)}, \bbX^{d}, \{\bbp_{\ell,\mathrm{o}}\}_{\ell=1}^{M} | \{\bbtheta_i\}_{i=1}^N)$. We solve the latter by updating $\{\bbtheta_i\}_{i=1}^{N}$ through policy gradient ascent. 

\begin{figure*}%
	\centering
	\begin{subfigure}{0.4\columnwidth}
		\includegraphics[width=1.3\linewidth, height = 0.95\linewidth, trim = {3cm 0cm 1cm 1cm}, clip]{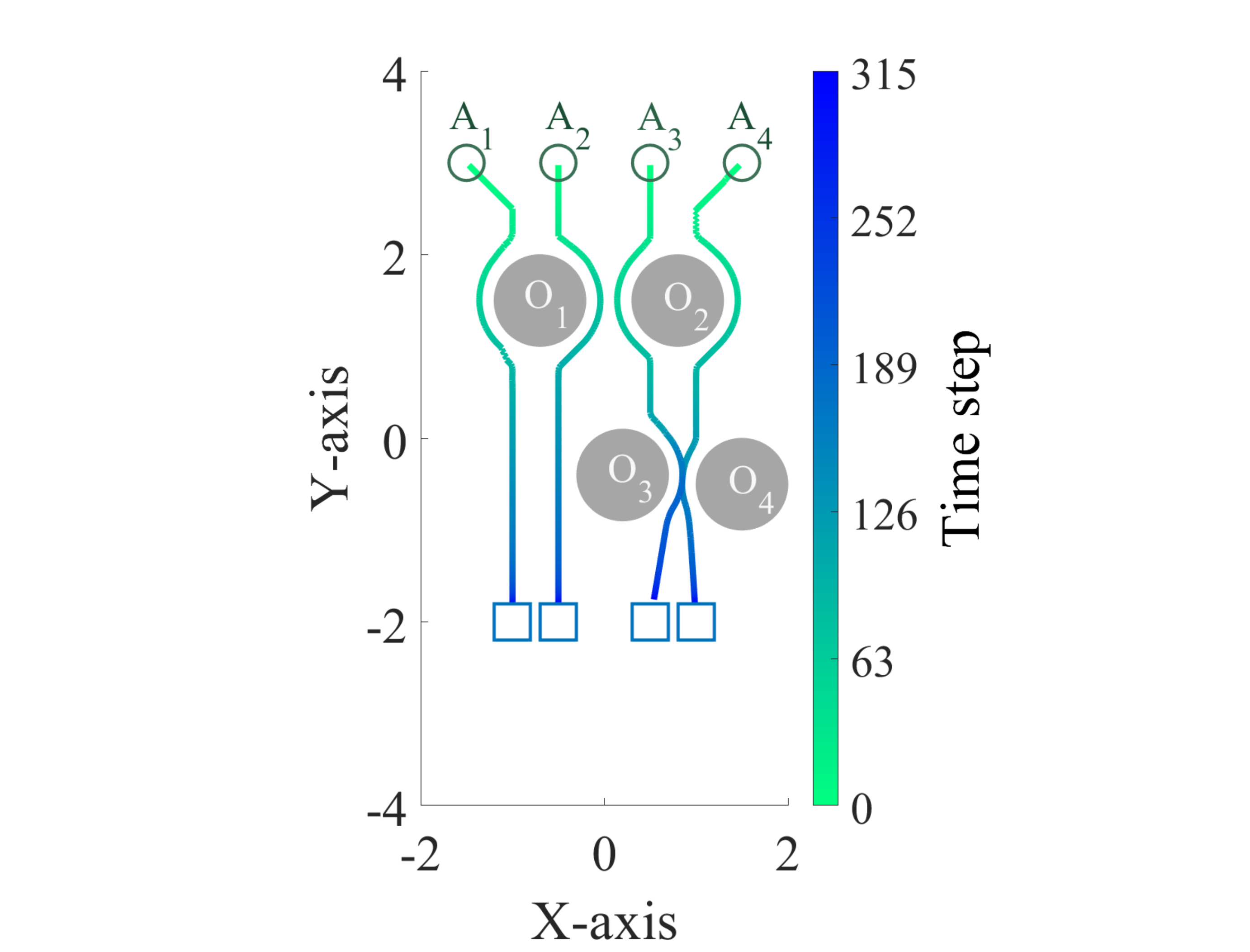}%
		\caption{}%
		\label{subfig2a}%
	\end{subfigure}\hfill\hfill%
	\begin{subfigure}{0.4\columnwidth}
		\includegraphics[width=1.1\linewidth,height = 0.95\linewidth]{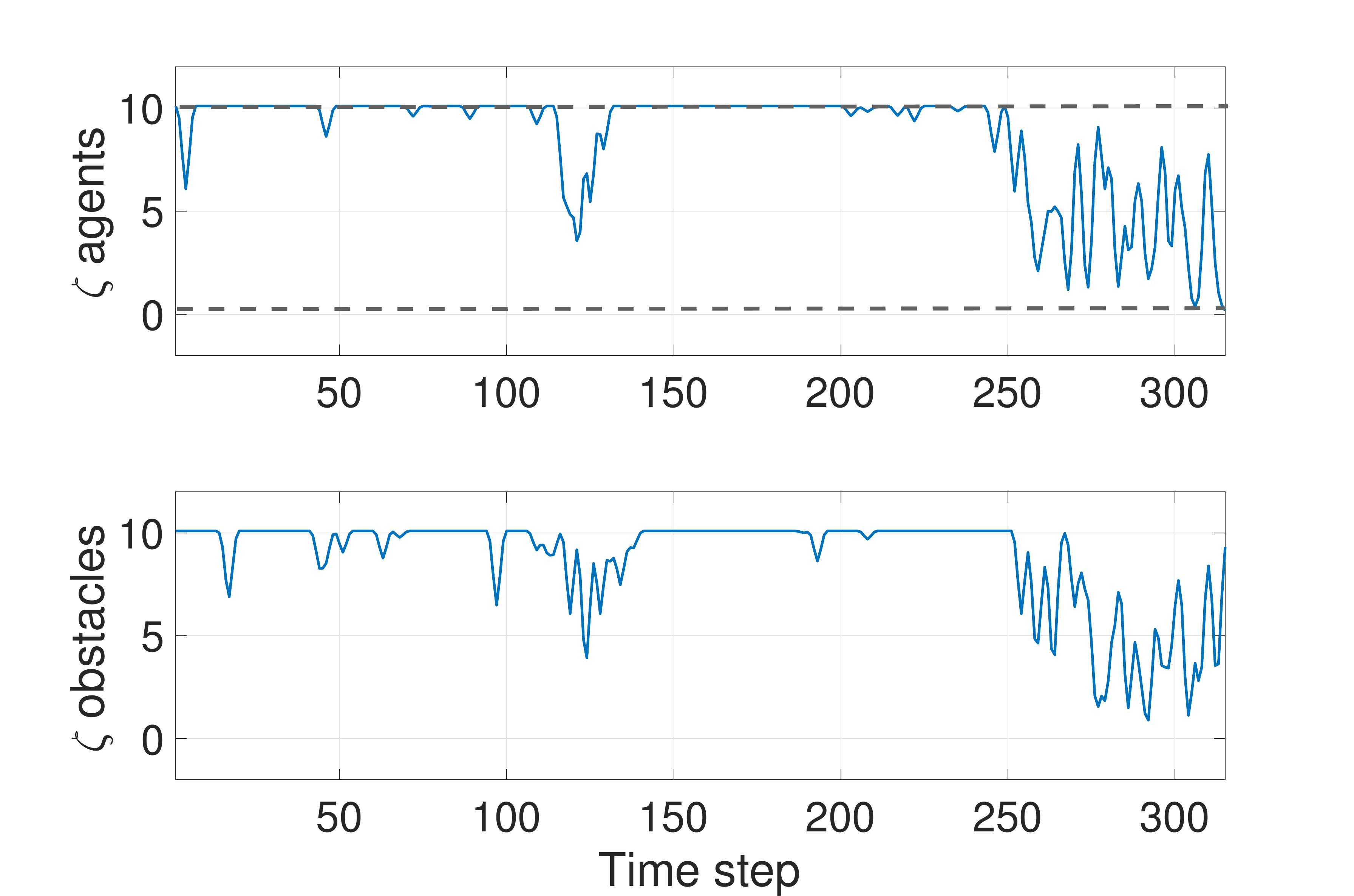}%
		\caption{}%
		\label{subfig2b}%
	\end{subfigure}\hfill\hfill%
	\begin{subfigure}{0.4\columnwidth}
		\includegraphics[width=1.1\linewidth, height = 0.95\linewidth]{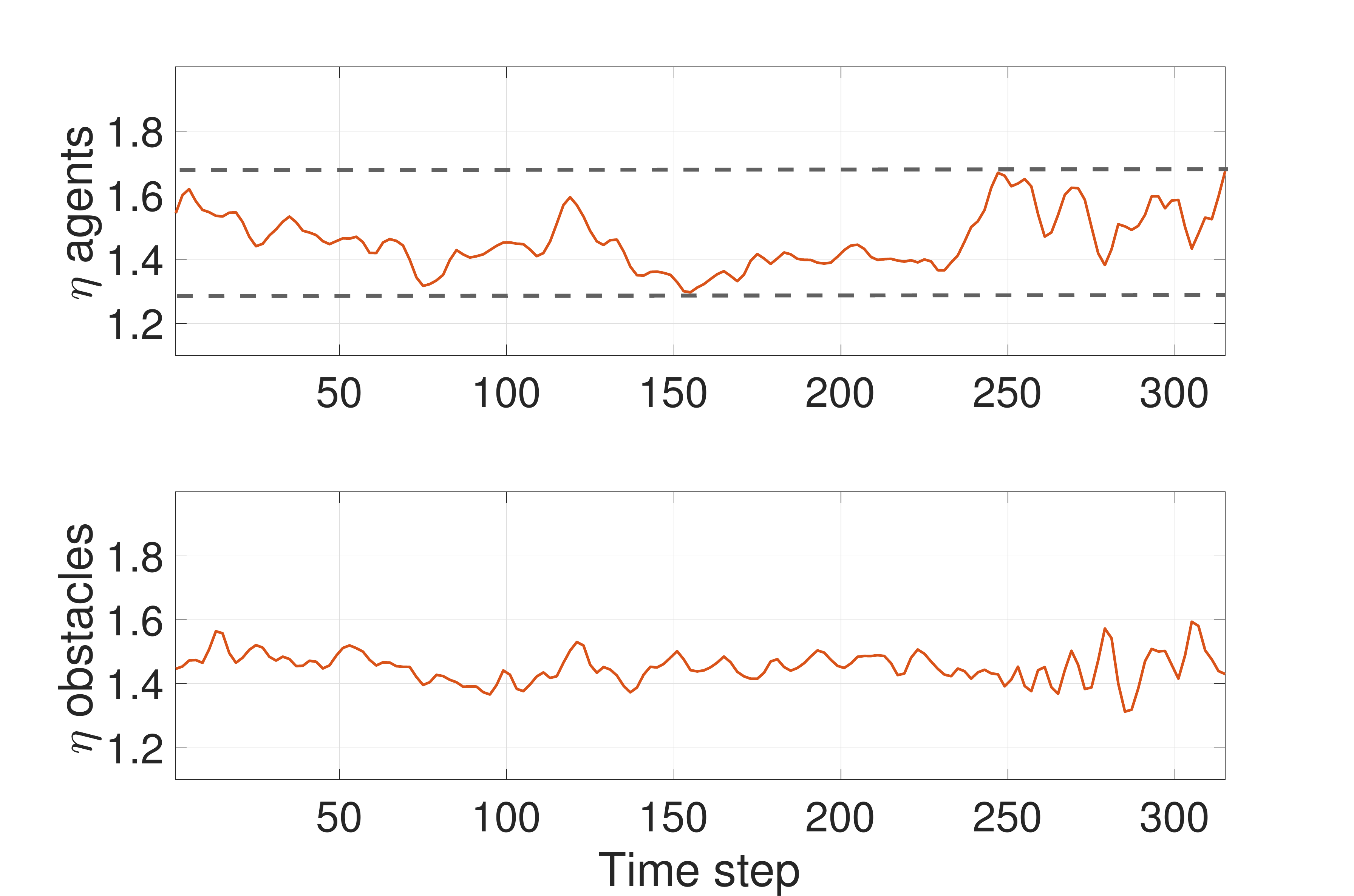}		
		\caption{}%
		\label{subfig2c}%
	\end{subfigure}
	\begin{subfigure}{0.4\columnwidth}
		\includegraphics[width=1.45\linewidth, height = 0.95\linewidth, trim = {7cm 0cm 0cm 1cm}, clip]{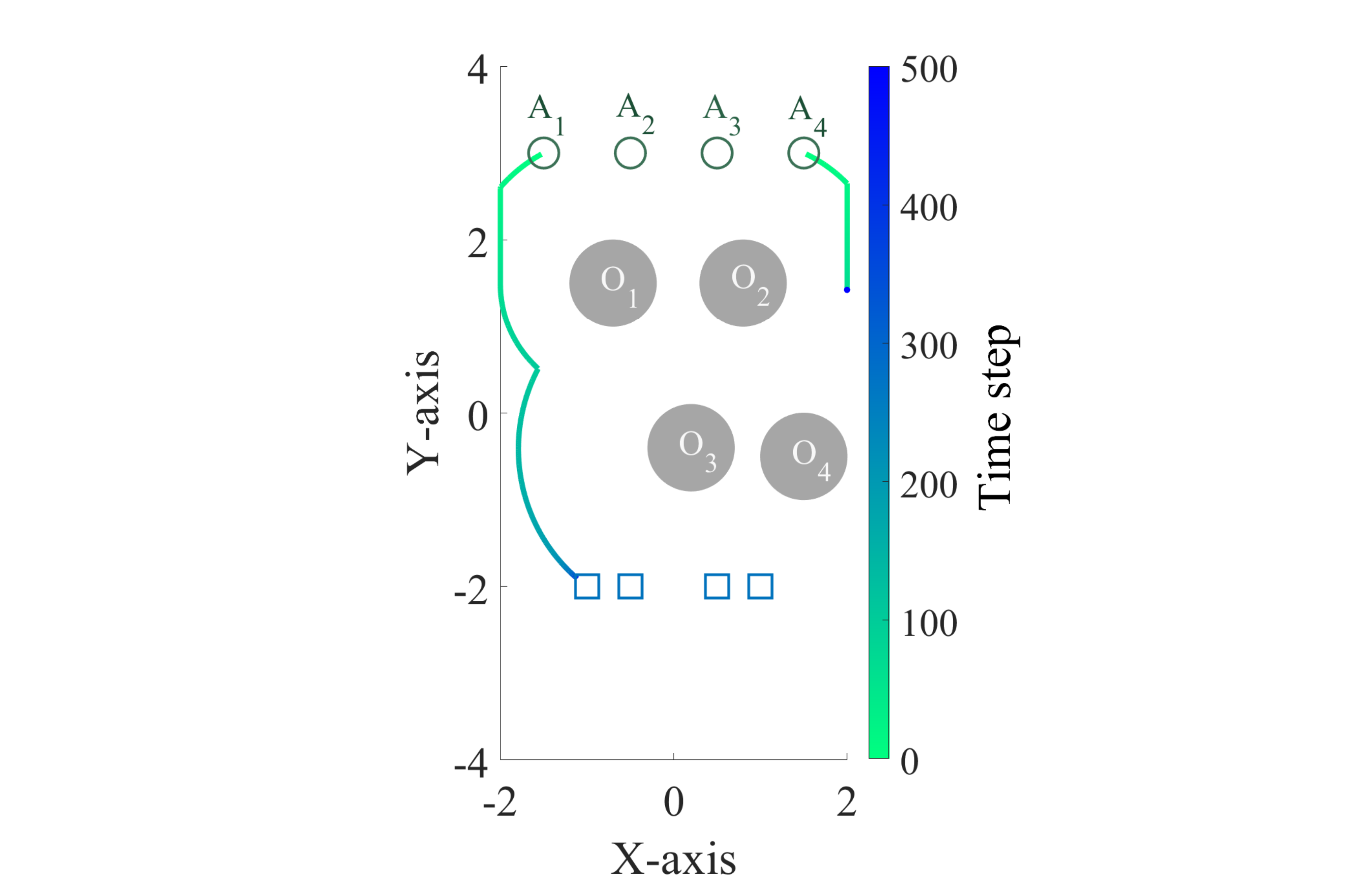}%
		\caption{}%
		\label{subfig3a}%
	\end{subfigure}\hfill
	\begin{subfigure}{0.4\columnwidth}
		\includegraphics[width=1.45\linewidth,height = 0.95\linewidth, trim = {7cm 0cm 0cm 1cm}, clip]{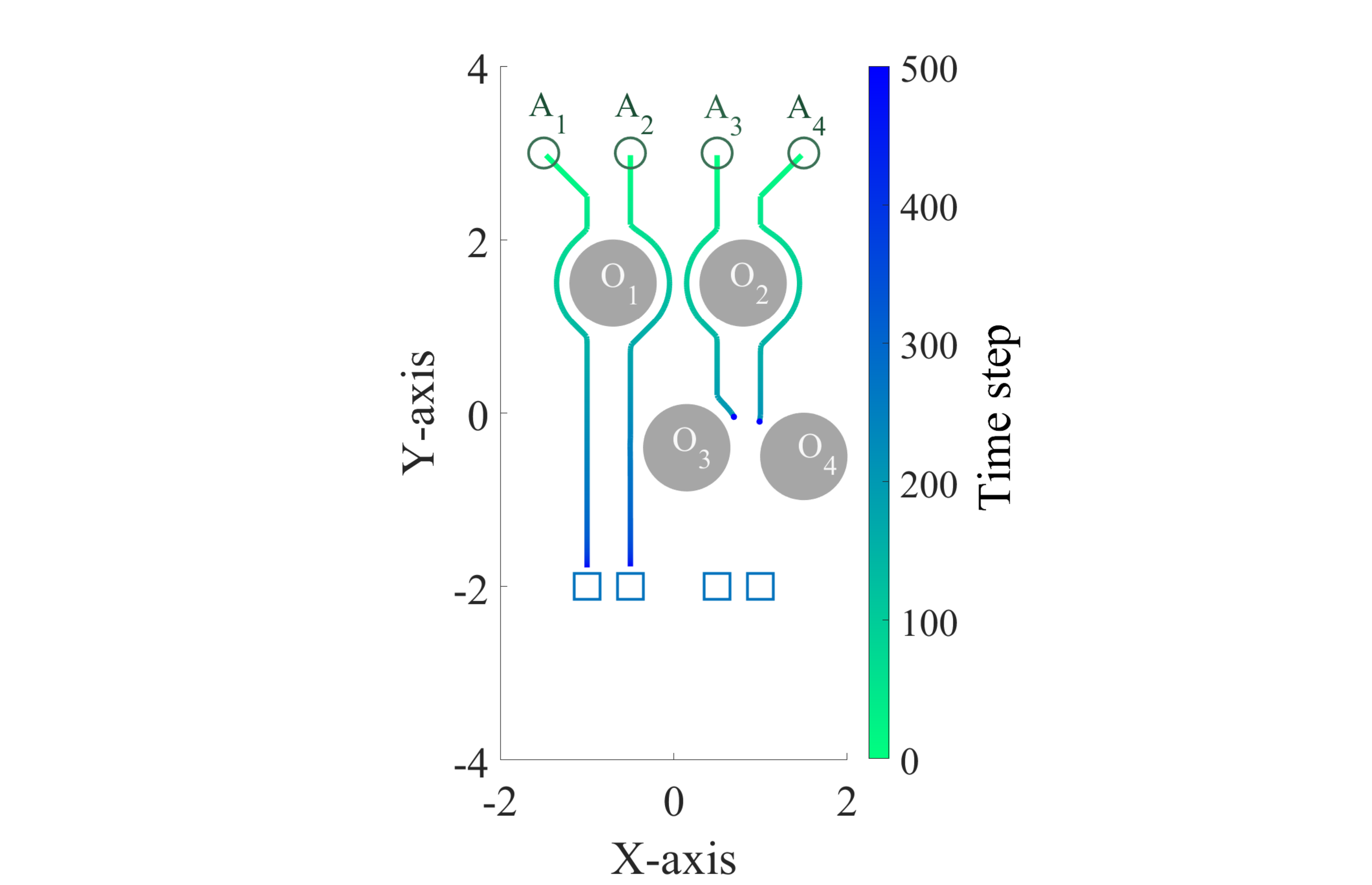}%
		\caption{}%
		\label{subfig3b}%
	\end{subfigure}
	\caption{\textbf{(a)} Agent trajectories with online CBFs generated by GNN-based policy. Green circles are initial positions, blue squares are goal positions, and grey circles are obstacles. Green-to-blue lines are agent trajectories and the color bar represents the time scale. \textbf{(b)-(c)} Time-varying CBF parameters $\zeta_{3,\rm a}$, $\zeta_{3,\rm o}$ and $\eta_{3,\rm a}$, $\eta_{3,\rm o}$ of agent $A_3$ w.r.t. the other agents and the obstacles. 
    The vertical lines in the top plots of (b)-(c) represent the maximal and minimal values of time-varying CBF parameters. 
    \textbf{(d)} Agent trajectories with the minimal fixed CBF parameters (i.e., the most conservative case). 
    The agents $A_2$, $A_3$ and $A_4$ have overly conservative CBFs and their controllers have no feasible solution. \textbf{(e)} Agent trajectories with the maximal fixed CBF parameters (i.e., the most aggressive case). 
    The agents $A_3$ and $A_4$ have overly aggressive CBFs and get stuck before the narrow passage between $O_3$ and $O_4$, where controllers have no feasible solution.
    }\label{fig2}\vspace{-6mm}
\end{figure*}

\subsection{Graph Neural Networks}

We  parameterize CBF-tuning policies with GNNs, which allow for decentralized execution. They are inherently \textit{permutation equivariant} (independent of agent ordering), and hence, generalize to unseen agent constellations~\cite{Scarselli2009, velivckovic2018graph, gao2021stochastic}. 

Motivated by the observation that CBFs need only relative information (e.g., relative positions between agents and obstacles) [cf. \eqref{eq:nu_CBF}], we design a \textit{translation-invariant} GNN that leverages message passing mechanisms to generate CBF parameters with {relative information}. For each agent $A_i$ with its local state $\bbx_i$, the states of neighboring agents $\{\bbx_j\}_{j \in \ccalN_i}$ and the positions of neighboring obstacles $\{\bbp_{\ell, \mathrm{o}}\}_{\ell \in \ccalN_i}$, it generates CBF parameters with the message aggregation functions $\ccalF_{\rm m, a}, \ccalF_{\rm m, o}$ and the feature update function $\ccalF_{\rm u}$ as 
\begin{align}\label{eq:GNN}
    (\bbzeta_i^{(t)}\!, \bbeta_i^{(t)}) &\!=\! \bbPhi_i(\bbx_i^{(t)}, \{\bbx_j^{(t)}\}_{j \in \ccalN_i},\{\bbp_{\ell,\mathrm{o}}\}_{\ell \in \ccalN_i}, \bbtheta_i)\\
    & \!=\! \ccalF_{\rm u}\Big(\! \sum_{j \in \ccalN_i}\! \ccalF_{\rm m, a} (\bbx_j \!-\! \bbx_i) \!+\!\! \sum_{\ell \in \ccalN_i}\! \ccalF_{\rm m, o} (\bbp_{\ell, \mathrm{o}} \!-\! \bbp_i) \!\Big), \nonumber
\end{align}
where $\bbtheta_i$ are the function parameters of $\ccalF_{\rm m,a}, \ccalF_{\rm m, o}$ and $\ccalF_{\rm u}$. By sharing $\ccalF_{\rm m, a}, \ccalF_{\rm m, o}$ and $\ccalF_{\rm u}$ over all agents, we have $\bbtheta_1=\cdots=\bbtheta_N$ and thus $\bbPhi_1=\cdots=\bbPhi_N$. 

The GNN-based policy has the following properties:
\begin{enumerate}
    \item {Decentralized execution:} The functions $\ccalF_{\rm m, a}, \ccalF_{\rm m, o}, \ccalF_{\rm u}$ require only neighborhood information and the policy can be executed in a decentralized manner. 
    \item Translation invariance: The policy uses relative information and is invariant to translations in $\mathbb{R}^2$.
    \item Permutation equivariance: The functions $\ccalF_{\rm m, a}, \ccalF_{\rm m, o}, \ccalF_{\rm u}$ are homogeneous and the policy is equivariant to permutations (i.e., agent reorderings).
\end{enumerate}

\section{EXPERIMENTS}
We evaluate our approach in this section. First, we conduct a proof of concept with four agents and four obstacles. Then, we show how our approach solves navigation scenarios that are infeasible with fixed CBFs. Next, we show the generalization of our approach in scenarios with more obstacles. Lastly, we report results from real-world experiments. 

\subsection{Proof of Concept}

We consider an environment shown in Fig. \ref{subfig2a}. The agents have radius $0.15$m, and are initialized randomly in the top region of the workspace and tasked towards goal positions in the bottom region. The obstacles are of radius $0.5$m, and are distributed between the initial and goal positions of agents. 

\noindent \textbf{Implementation details.} The agents are represented by positions $\{\bbp_{i}\}_{i=1}^N$ and velocities $\{\bbv_{i}\}_{i=1}^N$, and the obstacles 
by positions $\{\bbp_{\ell,\rm o}\}_{\ell=1}^{M}$. At each time step, each agent generates desired CBF parameters with its local policy based on neighborhood information, and feeds the latter into the QP controller to generate the feasible velocity towards its destination. An episode ends if all agents reach destinations or the episode times out. 
The sensing range, i.e., the communication radius, is $2$m, the maximal velocity is $0.5$m per time step in each direction, the maximal time step is $500$, and the time interval is $0.05$s. At time $t$, the reward is defined as 
\begin{align}
	r_{i}^{(t)} = \Big(\frac{\bbp_{i}^{(t)}-\bbd_i}{\|\bbp_{i}^{(t)}-\bbd_i\|_2} \cdot \frac{\bbv_{i}^{(t)}}{\|\bbv_{i}^{(t)}\|_2} \Big) \|\bbv_{i}^{(t)}\|_2 + r^{(t)}_{QP},
\end{align}
where the first term rewards fast movement towards the destination and the second term represents the infeasibility penalty of the QP controller. The message aggregation and feature update functions of the GNN are multi-layer perceptrons (MLPs), and the training is conducted with PPO \cite{schulman2017proximal}.

\begin{figure*}%
	\centering
	\begin{subfigure}{0.33\columnwidth}
		\includegraphics[width=1.45\linewidth, height = 1.0\linewidth, trim = {8cm 0cm 0cm 1cm}, clip]{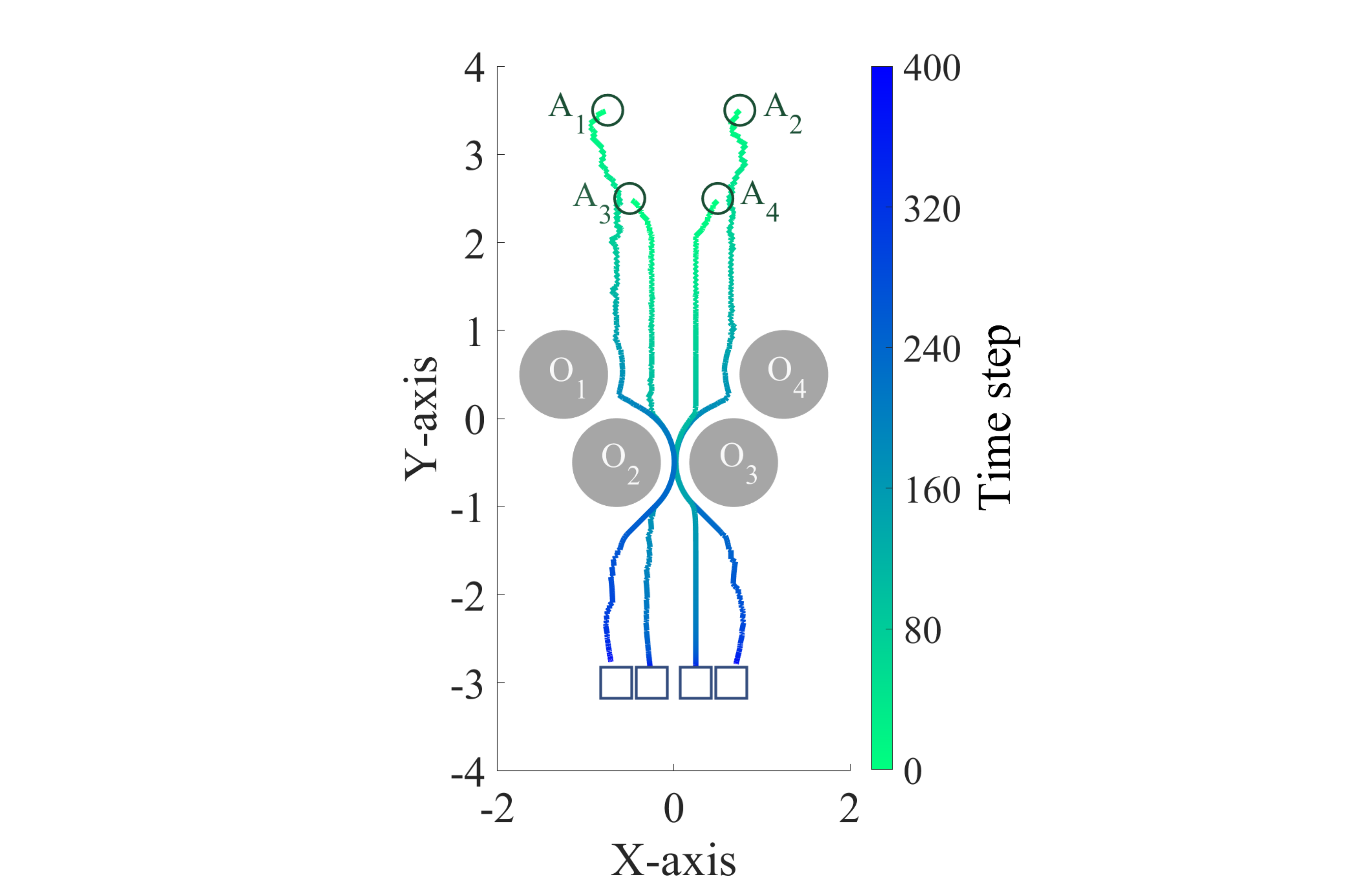}%
		\caption{}%
		\label{subfig4a}%
	\end{subfigure}
	\begin{subfigure}{0.33\columnwidth}
		\includegraphics[width=1.45\linewidth, height = 1.0\linewidth, trim = {8cm 0cm 0cm 1cm}, clip]{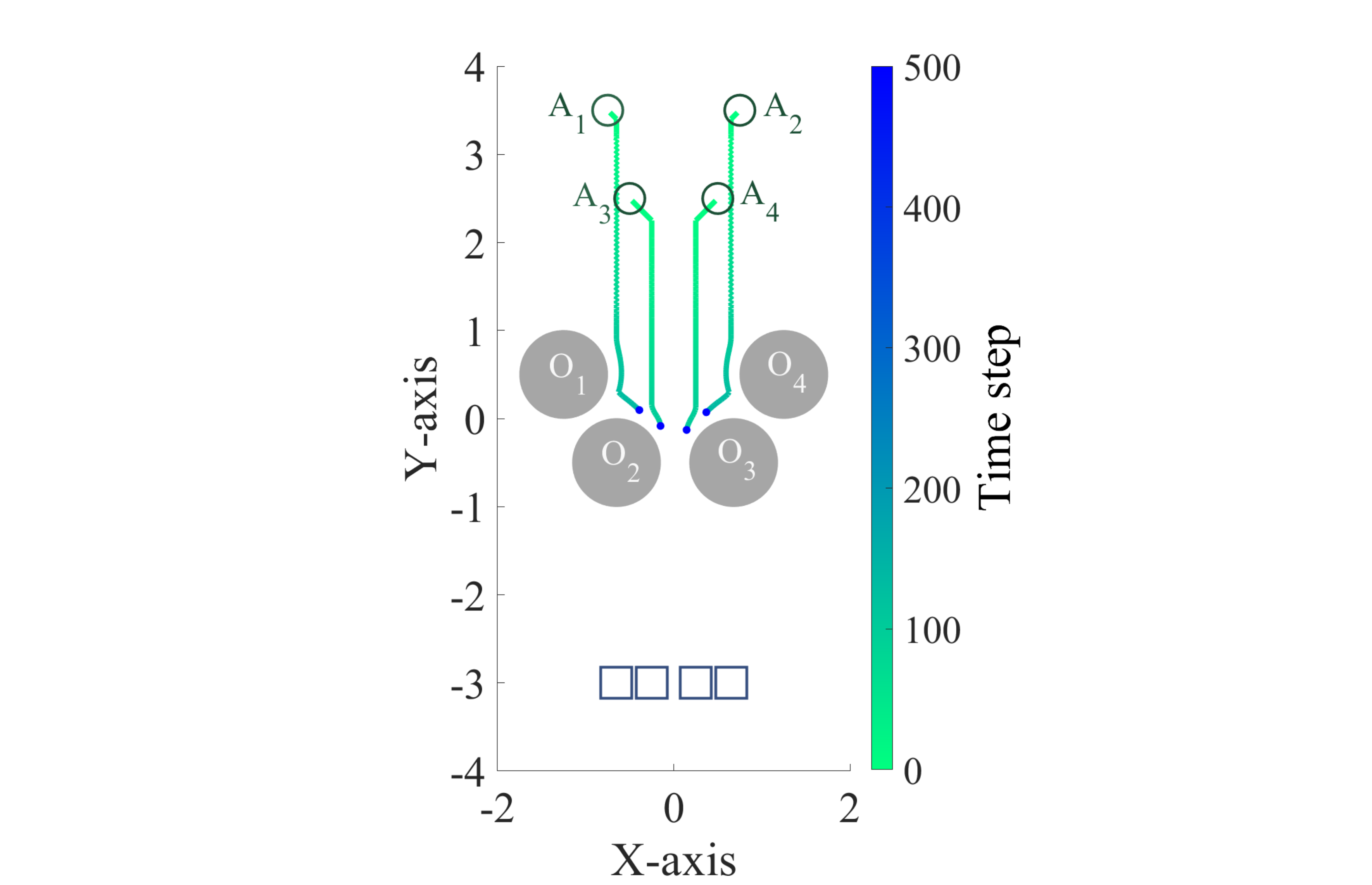}%
		\caption{}%
		\label{subfig4b}%
	\end{subfigure}
	\begin{subfigure}{0.33\columnwidth}
		\includegraphics[width=1.45\linewidth, height = 1.0\linewidth, trim = {8cm 0cm 0cm 1cm}, clip]{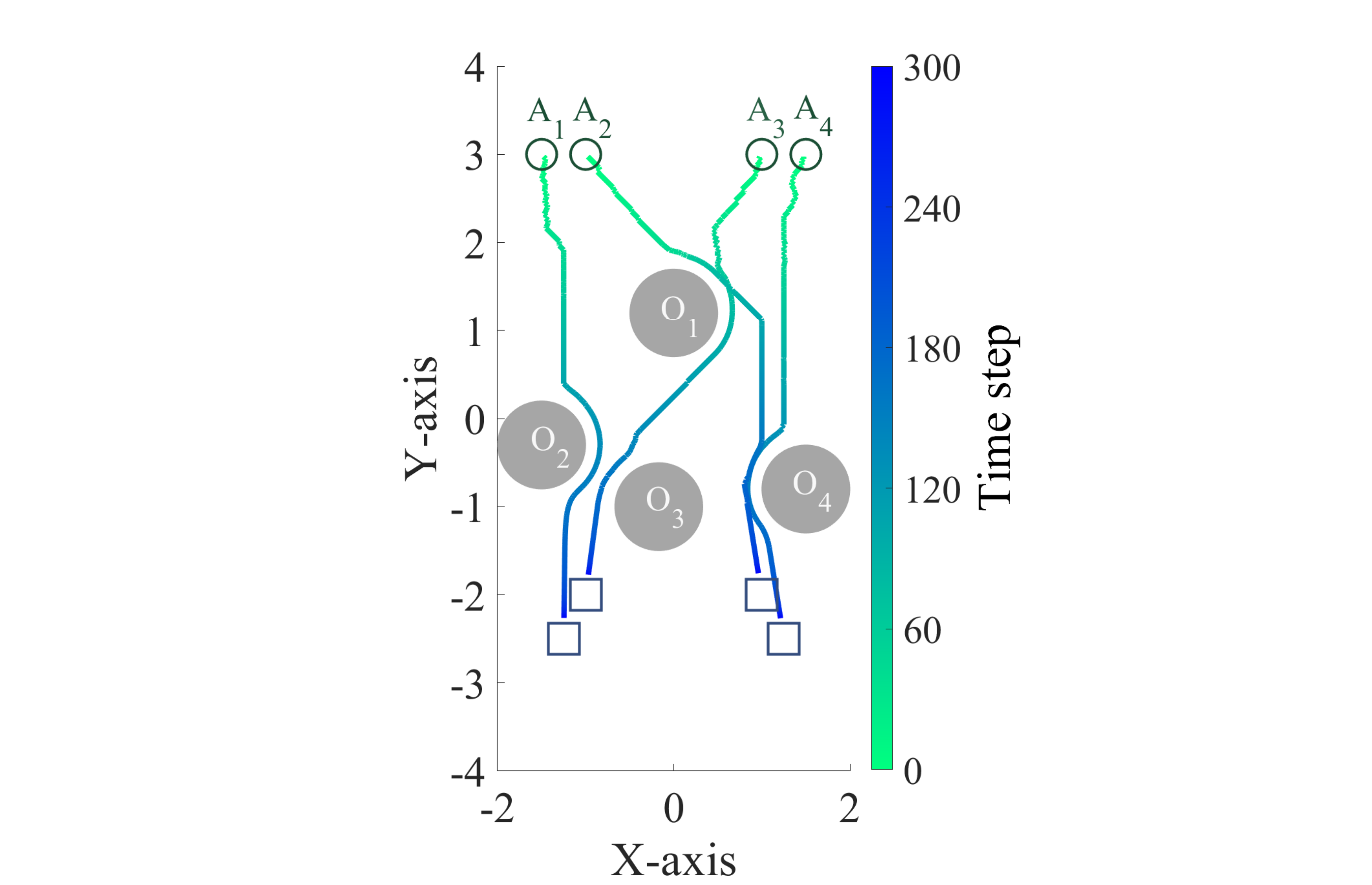}%
		\caption{}%
		\label{subfig4c}%
	\end{subfigure}
	\begin{subfigure}{0.33\columnwidth}
		\includegraphics[width=1.45\linewidth, height = 1.0\linewidth, trim = {8cm 0cm 0cm 1cm}, clip]{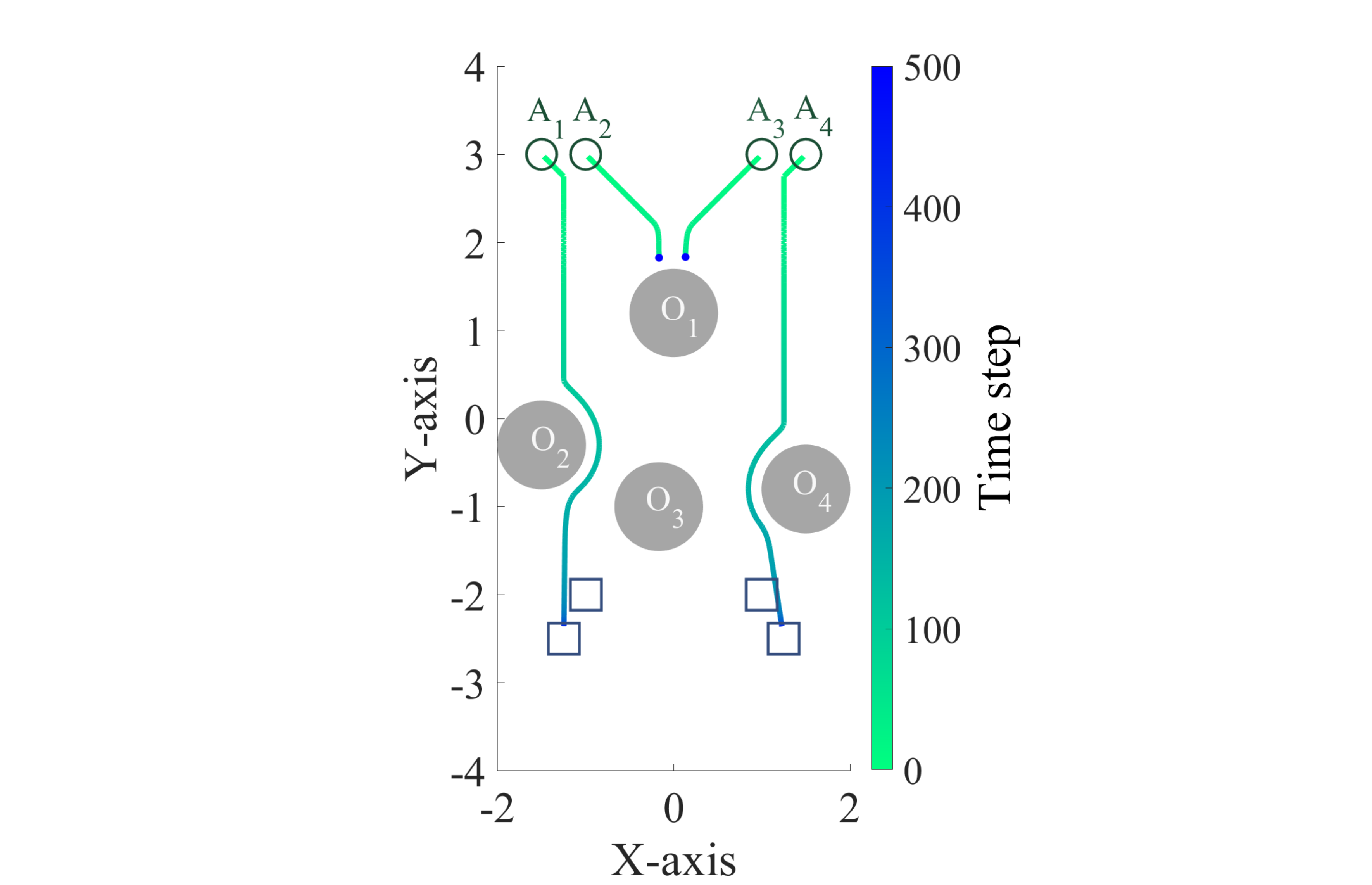}%
		\caption{}%
		\label{subfig4d}%
	\end{subfigure}
	\begin{subfigure}{0.33\columnwidth}
		\includegraphics[width=1.45\linewidth, height = 1.0\linewidth, trim = {8cm 0cm 0cm 1cm}, clip]{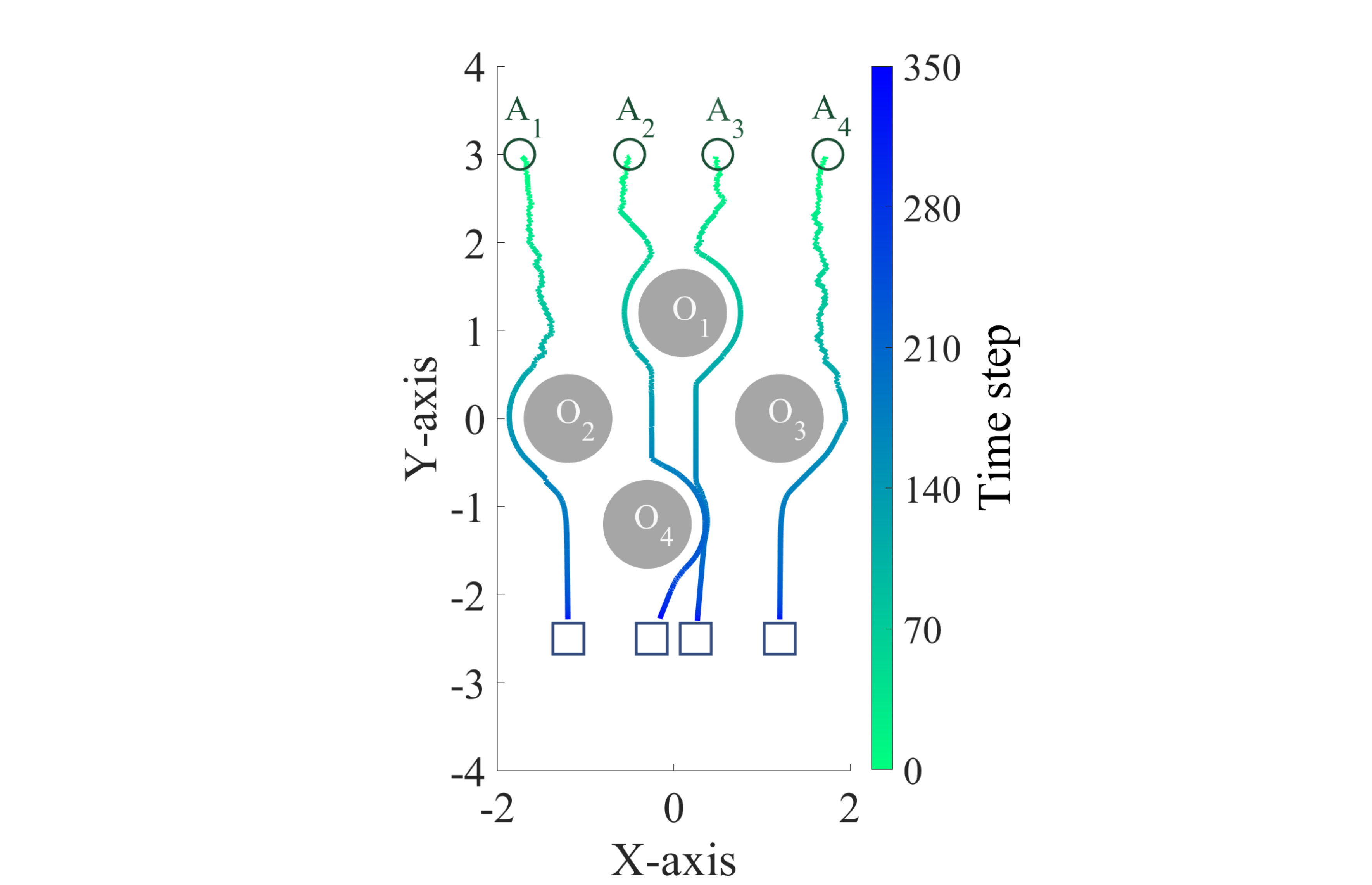}%
		\caption{}%
		\label{subfig4e}%
	\end{subfigure}
	\begin{subfigure}{0.33\columnwidth}
		\includegraphics[width=1.45\linewidth, height = 1.0\linewidth, trim = {8cm 0cm 0cm 1cm}, clip]{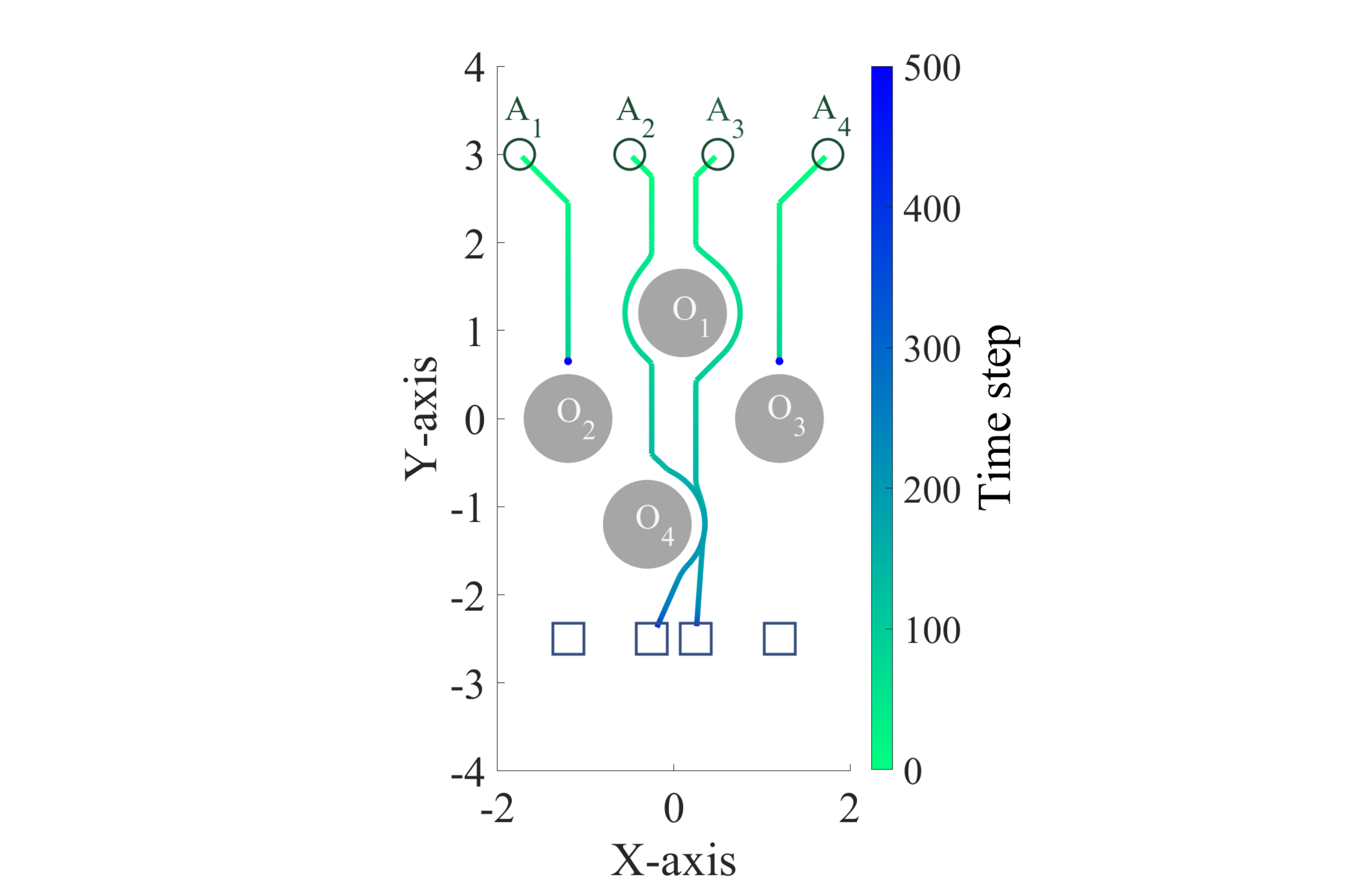}%
		\caption{}%
		\label{subfig4f}%
	\end{subfigure}%
	\caption{Agent trajectories with online CBFs generated by GNN-based tuning policy and optimal fixed CBFs selected by exhaustive grid-search in different infeasible scenarios. \textbf{(a)} \textit{Narrow Passage} scenario with online CBFs. \textbf{(b)} \textit{Narrow Passage} scenario with optimal fixed CBFs. \textbf{(c)} \textit{Cross} scenario with online CBFs. \textbf{(d)} \textit{Cross} scenario with optimal fixed CBFs. \textbf{(e)} \textit{Singularity} scenario with online CBFs. \textbf{(f)} \textit{Singularity} scenario with optimal fixed CBFs.}\label{fig4}\vspace{-6mm}
\end{figure*}

\noindent \textbf{Performance.} Fig. \ref{subfig2a} shows the agent trajectories with online CBFs. The agents move smoothly from initial positions to destinations without collision. Figs. \ref{subfig2b}-\ref{subfig2c} show the variation of CBF parameters $\bbzeta$, $\bbeta$ of an example agent $A_3$ w.r.t. the other agents and obstacles, respectively. We see that \textit{(i)} the values of $\bbzeta$ 
remain maximal for the majority of its trajectory, which can be interpreted as a relaxation of CBF constraints to achieve fast velocities towards destination; \textit{(ii)}, the values of $\bbzeta$ drop and the values of $\bbeta$ increase between time step $100$ and $150$, which render CBF constraints conservative to avoid 
inter-agent congestion. I.e., $A_3$ slows its velocity 
for $A_4$ when preparing to pass through the narrow passage between $O_3$ and $O_4$; \textit{(iii)}, the values of $\bbzeta$ and $\bbeta$ tend to be random after time step $250$, because $A_3$ is close to its destination and CBF parameters play little role 
at that stage.

To show the trade-off between conservative and aggressive behavior inherent in CBFs, we select the minimal and maximal values from time-varying CBF parameters in Figs. \ref{subfig2b}-\ref{subfig2c} (dashed lines), corresponding to the most conservative and aggressive CBFs, and perform navigation for fixed CBFs \cite{srinivasan2018control,dawson2022safe} with these selected parameters. For the minimal values in Fig. \ref{subfig3a}, $A_2$, $A_3$ keep still and $A_1$, $A_4$ move along the environment boundary, with overly conservative trajectories, and only $A_1$ reaches its destination. For the maximal values in Fig. \ref{subfig3b}, while all agents aggressively move towards destinations, $A_3$ and $A_4$ get stuck before the narrow passage between $O_3$ and $O_4$. This is because the agents are too close to each other and the obstacles, where their controller has no feasible solution. This highlights the importance of our approach, which provides online CBFs. 

\subsection{Feasibility}

We perform our approach in three distinct navigation scenarios: \textit{Narrow Passage}, \textit{Cross}, and \textit{Singularity}. For the fixed CBFs, we employ an exhaustive grid-search to find optimal parameters $(\zeta, \eta)$ from $[0.1,10] \times [1.0, 2.0]$ for $100$ combinations. 

Fig. \ref{fig4} shows the performance of our approach and the optimal fixed CBFs in our three scenarios. Overall, our results show that, while we exhaustively traversed the parameter space, there still exist scenarios where there is no solution for fixed CBFs. In contrast, the online CBFs can solve these infeasible scenarios (i.e., all agents reach their destinations successfully). For the \textit{Narrow Passage} scenario in Figs. \ref{subfig4a}-\ref{subfig4b} and the \textit{Cross} scenario in Figs. \ref{subfig4c}-\ref{subfig4d}, our approach deconflicts agents by prioritizing them with varying degrees of conservative / aggressive behaviors. For the \textit{Singularity} scenario in Figs. \ref{subfig4e}-\ref{subfig4f}, the agent and its destination are aligned with an obstacle in the middle. The fixed CBF-based controller generates controls on the edge or vertex of the admissible control set, and hence, agents get stuck in local minima \cite{wang2018multi}, while our approach helps agents escape such conditions due to online CBF tuning.

\subsection{Generalization}\label{subsec:case2}

\begin{figure}%
	\centering
	\begin{subfigure}{0.49\columnwidth}
		\includegraphics[width=1\linewidth,height = 0.8\linewidth, trim = {4cm 0cm 3cm 1cm}, clip]{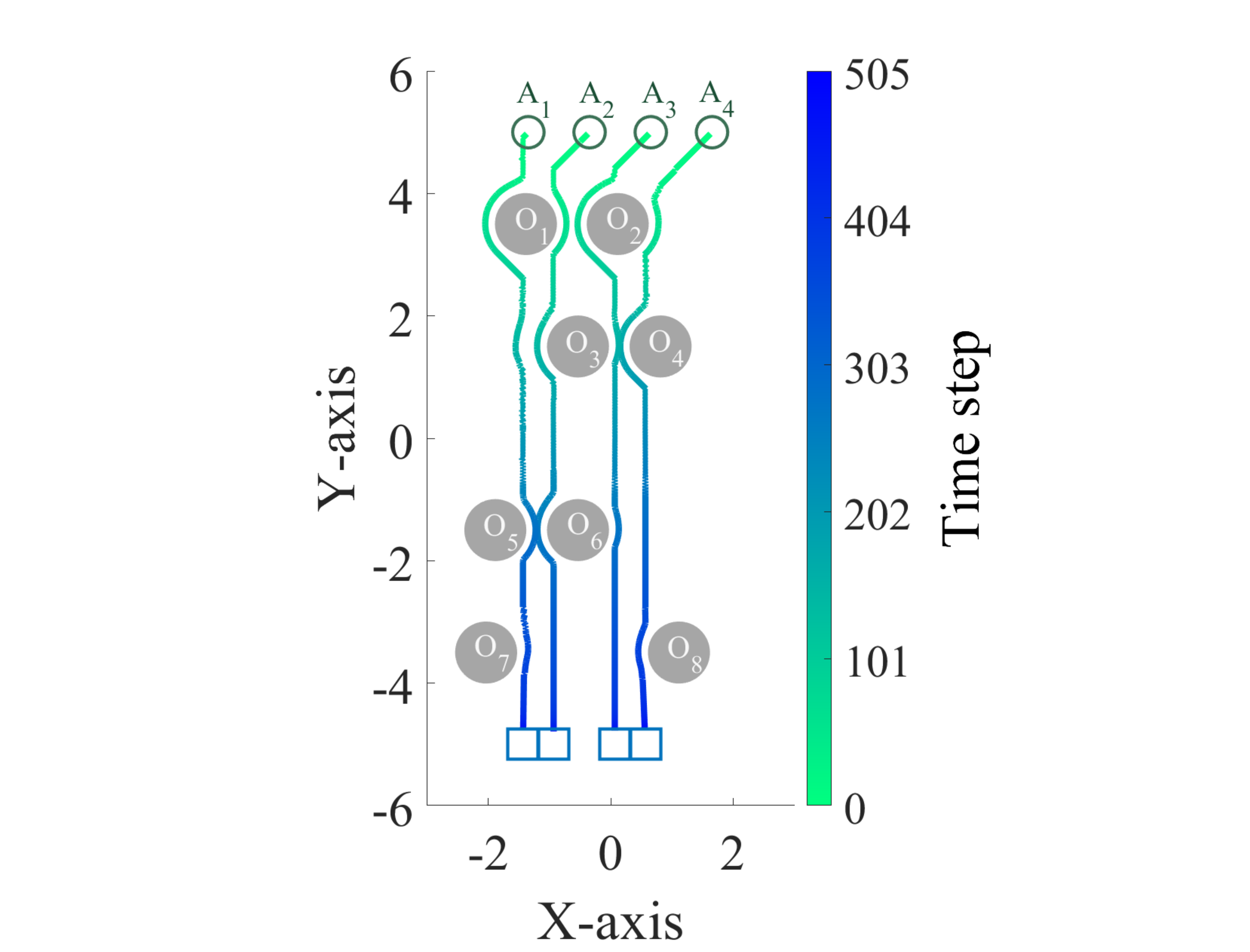}%
		\caption{}%
		\label{subfig5a}%
	\end{subfigure}
	\begin{subfigure}{0.49\columnwidth}
		\includegraphics[width=1\linewidth,height = 0.8\linewidth, trim = {2cm 1cm 1cm 1cm}]{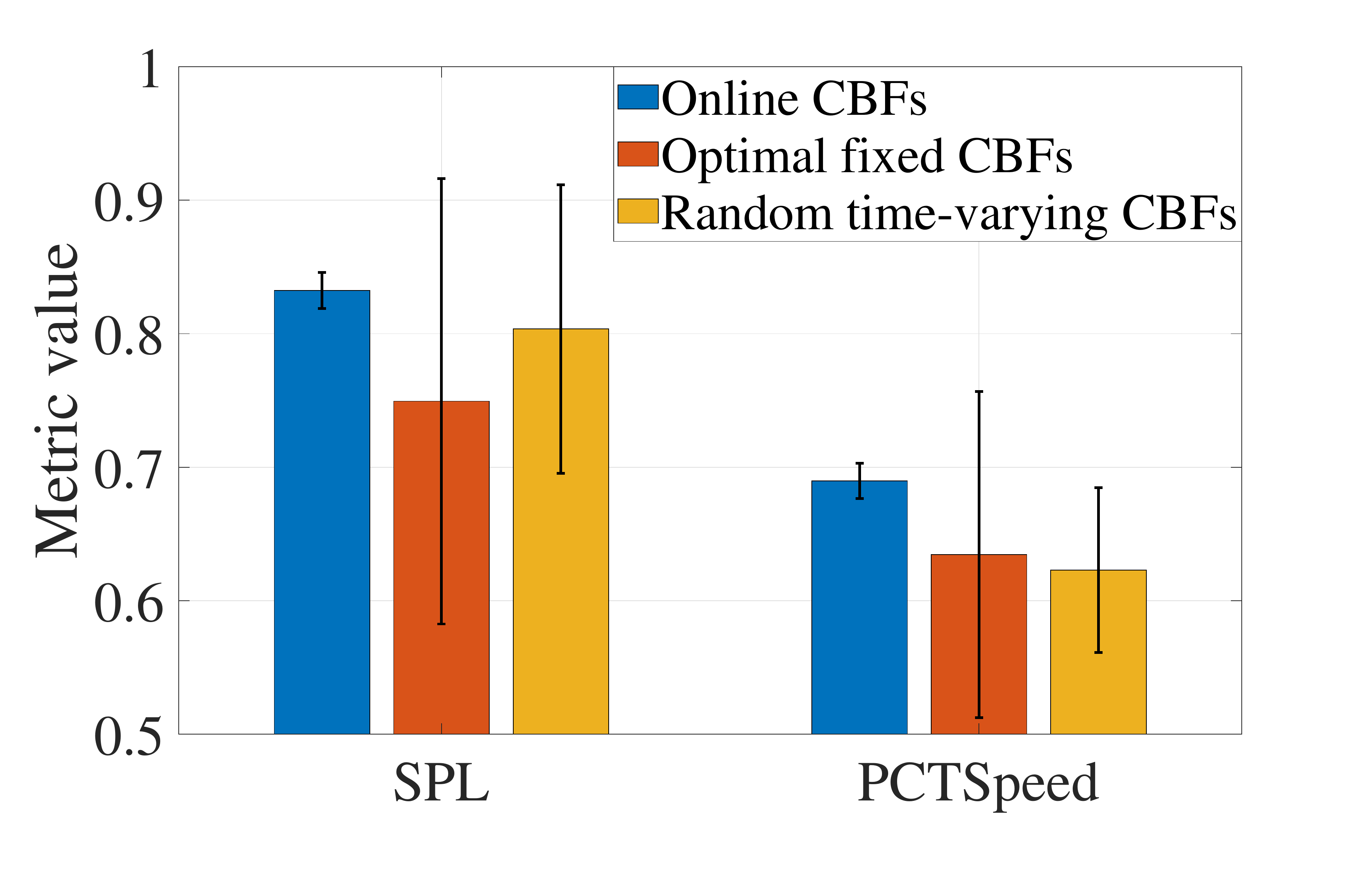}%
		\caption{}%
		\label{subfig5b}%
	\end{subfigure}
	\caption{\textbf{(a)} Environment used to train GNN-based policy for online CBFs and to conduct grid-search for fixed CBFs. Agent trajectories are generated with online CBFs of the trained GNN-based policy. \textbf{(b)} Performance comparison between online CBFs generated by GNN-based policy, optimal fixed CBFs with exhaustive grid-search and time-varying CBFs with random parameters.}\label{fig5}\vspace{-6mm}
\end{figure}

We show the generalization of our approach by testing the trained policy on previously unseen environments. We consider two baselines: \emph{(i)} optimal fixed CBFs with exhaustive grid-search and \emph{(ii)} time-varying CBFs with random parameters. The first searches the parameter space exhaustively and selects the optimal values for fixed CBFs. As existing works primarily concentrate on identifying the optimal fixed CBFs for specific tasks \cite{xiao2019explicit, 9303857, usevitch2021adversarial}, we consider this baseline to approximate the optimal performance of controllers with fixed CBFs. Note that it is inefficient and included here solely for reference. The second selects random CBF parameters every $10$ time steps, which is as efficient as our approach. 

We consider larger environments with $8$ obstacles, where the maximal time step is $750$. We train our approach for online CBFs and conduct grid-search for fixed CBFs in the environment shown in Fig. \ref{subfig5a}, and test them by randomly shifting initial, goal and obstacle positions. The performance is measured by two metrics: \emph{(i)} Success weighted by Path Length (SPL) \cite{anderson2018evaluation} and \emph{(ii)} the percentage to the maximal speed (PCTSpeed). The former is a stringent measure combining the success rate and the path length, while the latter represents the ratio of the average speed to the maximal one. 

Fig. \ref{subfig5b} shows the results averaged over $20$ random initializations. Our approach outperforms the baselines in both metrics, with higher expectations and lower standard deviations. This corresponds to theoretical findings that \emph{(i)} online CBFs coordinate agents' conservative / aggressive behaviors based on instantaneous environment states, which allows smooth navigation without congestion for higher expected performance; \emph{(ii)}, online CBFs deconflict agents to solve infeasible scenarios, which improves robustness for lower standard deviations. Random CBFs exhibit a higher SPL than fixed CBFs because they deconflict some infeasible scenarios by randomly prioritizing agents and have a higher success rate. However, random CBFs show a lower PCTSpeed because randomly coordinating agents exhibits poor performance, even though navigation tasks are successful. 

\subsection{Real-World Experiments}
\begin{figure}[t]
\centering
\includegraphics[width=0.4\textwidth]{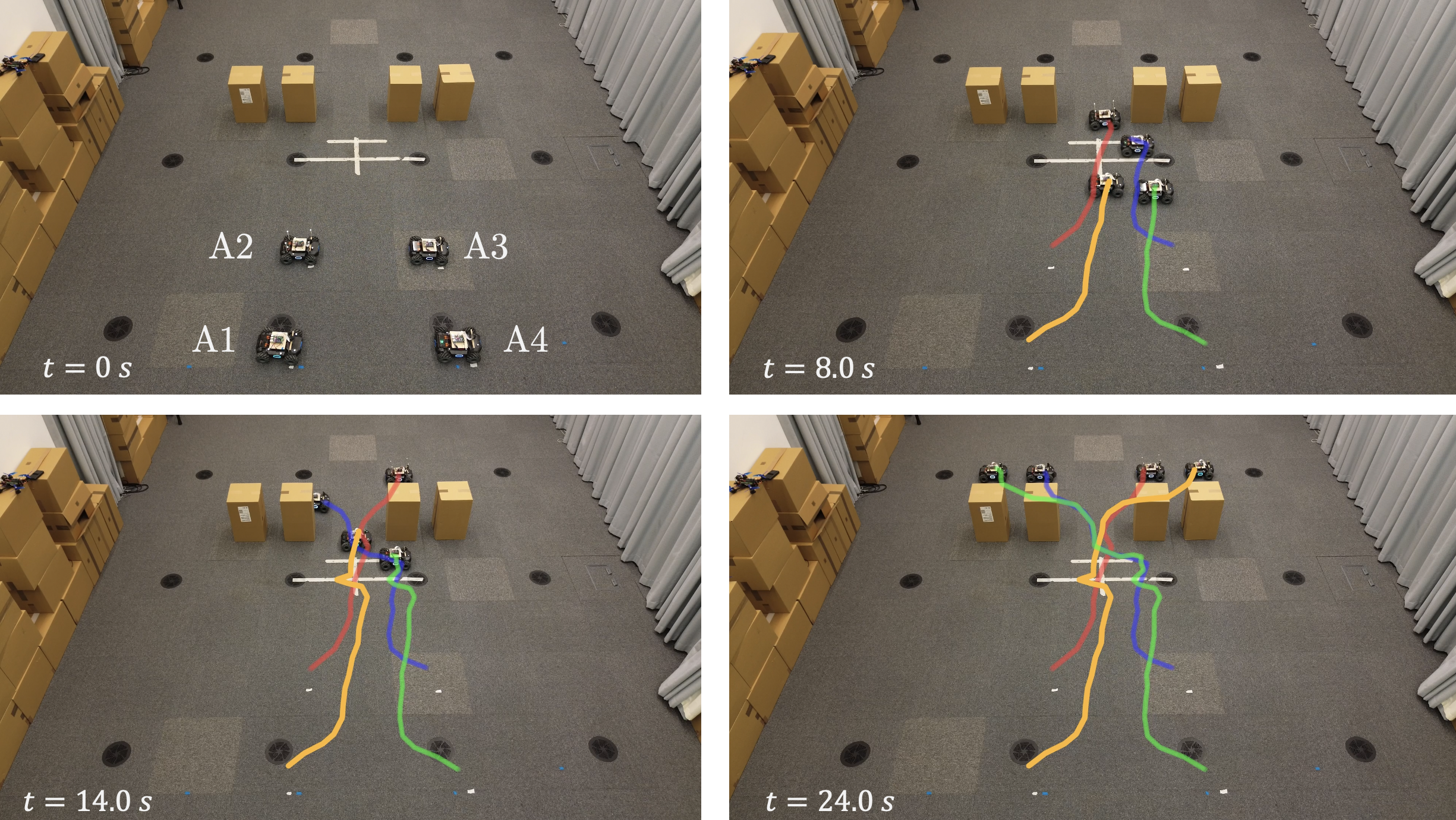}
\captionof{figure}{Real-world experiments with DJI's Robomasters. Robots are required to pass through a narrow passage and reach predefined goal positions. The online CBFs are able to deconflict robots.
}
\label{fig:exp}\vspace{-6mm}
\end{figure}
We conduct real-world experiments to validate our approach. We consider a narrow passage scenario that requires deconfliction for multi-robot navigation. We use four customized DJI Robomasters with Raspberry Pi. 
Each robot has a partially observable space with a sensing range of $2$m, and employs an external telemetry (OptiTrack) for localization. We use ROS2 as communication middle-ware. At each time, the robot receives its current state and the neighbors' states, and deploys the decentralized controller for navigation. 

Fig. \ref{fig:exp} shows that our approach steers all robots to their destinations without collision, because online CBFs deconflict robots by adapting between conservative and aggressive constraint values. When performing the same experiment with (optimized) fixed CBFs, robots fail to navigate through the narrow passage because online deconfliction is not facilitated. This insight corroborates our theoretical analysis and numerical simulations \cite[Video Link]{video}. 

\section{CONCLUSION}
This paper proposed \emph{online} CBFs for decentralized multi-agent navigation. We formulated the problem of multi-agent navigation as a quadratic programming with CLFs for state convergence and CBFs for safety constraints, and proposed an online CBF optimization that tunes CBFs based on instantaneous state information in a dynamic environment. We solved this problem by leveraging RL with GNN parameterization. The former allows for model-free training and the latter provides decentralized agent controllers. We show, through simulations and real-world experiments, that our approach coordinates agent behaviors to deconflict their trajectories and improve overall navigation performance, all the while ensuring safety. In future work, we will extend our work on higher-dimensional non-linear systems.







{\small
	\bibliographystyle{IEEEtran}
	\bibliography{myIEEEabrv,biblioOp,AP_bib}
}




\end{document}

%% file: mySections.tex
\usepackage{needspace}

